\documentclass[]{xiaomiev}
\usepackage[page,header]{appendix}
\usepackage{booktabs}
\usepackage{multirow}
\usepackage{graphicx}
\usepackage{hyperref}
\usepackage{amsmath}
\usepackage{amssymb}
\usepackage[colorinlistoftodos]{todonotes}
\usepackage{float}
\usepackage{subcaption}
\usepackage[numbers]{natbib}
\usepackage{enumitem}
\usepackage{listings}
\usepackage[ruled,linesnumbered]{algorithm2e}
\setlist[itemize]{leftmargin=15pt}

\definecolor{hxOrange}{HTML}{ED722E}
\definecolor{hxNavy}{HTML}{1B262C}
\definecolor{hxBlue}{HTML}{0F4C75}
\definecolor{hxTeal}{HTML}{3282B8}
\definecolor{hxSky}{HTML}{BBE1FA}
\definecolor{hxKey}{HTML}{0F4C75}     
\definecolor{hxStr}{HTML}{8A5A00}     
\definecolor{hxComment}{HTML}{6B7B8C} 
\lstdefinelanguage{yaml}{
  sensitive=true,
  morecomment=[l]{\#},
  morestring=[b]",
  morestring=[b]',
  commentstyle=\color{hxComment}\itshape,
  stringstyle=\color{hxStr},
  keywords={true,false,null},
  keywordstyle=\color{hxTeal}\bfseries,
  moredelim=**[il][\color{hxKey}\bfseries]{:},
  moredelim=[l][\color{hxTeal}]{-\ },
}
\newtcblisting{promptfile}[2][]{%
  breakable, listing only,
  colback=hxSky!10, colframe=hxNavy, colbacktitle=hxNavy, coltitle=white,
  boxrule=0.6pt, left=4pt, right=4pt, top=3pt, bottom=3pt,
  title={\scriptsize\textbf{\texttt{#2}}}, #1}
\newtcblisting{manifestfile}[2][]{%
  breakable, listing only,
  listing options={language=yaml},
  colback=hxTeal!8, colframe=hxBlue, colbacktitle=hxBlue, coltitle=white,
  boxrule=0.6pt, left=4pt, right=4pt, top=3pt, bottom=3pt,
  title={\scriptsize\textbf{\texttt{#2}}}, #1}
\newcommand{\mfsec}[1]{\par\smallskip\textbf{\textcolor{hxBlue}{#1}}%
  \par\vspace{1pt}\noindent\rule{\linewidth}{0.3pt}\par\vspace{2pt}\noindent}
\newtcolorbox{manifestcard}[1]{%
  colback=hxTeal!8, colframe=hxBlue, colbacktitle=hxBlue, coltitle=white,
  fonttitle=\scriptsize\bfseries, fontupper=\footnotesize,
  boxrule=0.6pt, left=5pt, right=5pt, top=4pt, bottom=5pt,
  before upper={\setlength{\parindent}{0pt}\raggedright},
  title={#1}}
  
\RequirePackage{xspace}
\makeatletter
\DeclareRobustCommand\onedot{\futurelet\@let@token\@onedot}
\def\@onedot{\ifx\@let@token.\else.\null\fi\xspace}

\makeatother

\newcommand{\ProjectNameTitle}{Harness{\textcolor[RGB]{237,114,46}{X}}\xspace}

\newcommand{\ProjectName}{HarnessX\xspace}

\title{\ProjectNameTitle: A Composable, Adaptive, and Evolvable Agent Harness Foundry}

\author{Darwin Agent Team}

\contribution{See \hyperref[sec:contributions]{Contributions and Acknowledgments} section for a full author list.}

\abstract{
AI agent performance depends critically on the runtime harness, comprising the prompts, tools, memory, and control flow that mediate how a model observes, reasons, and acts. Yet today's harnesses remain largely hand-crafted and static: each new model or task still demands bespoke scaffolding, and the rich traces produced during execution are rarely distilled back into systematic improvement. We introduce \textbf{\ProjectName}, a foundry for composable, adaptive, and evolvable agent harnesses. \ProjectName assembles typed harness primitives via a substitution algebra, adapts them through AEGIS, a trace-driven multi-agent evolution engine grounded in an operational mirror between symbolic adaptation and reinforcement learning, and closes the harness--model loop by turning trajectories into both harness updates and model training signal. Across five benchmarks (ALFWorld, GAIA, WebShop, $\tau^3$-Bench, and SWE-bench Verified), \ProjectName yields an average gain of +14.5\% (up to +44.0\%), with gains largest where baselines are lowest. 
These results suggest that agent progress need not come from model scaling alone: composing and evolving runtime interfaces from execution feedback is an actionable and complementary lever. 
Project homepage: \textcolor{hxOrange}{\href{https://darwin-agent.github.io/HarnessX/}{https://darwin-agent.github.io/HarnessX/}.}
  
}

\begin{document}
\maketitle
\vspace{-10pt}
{%
  \centering
  \includegraphics[width=0.8\textwidth]{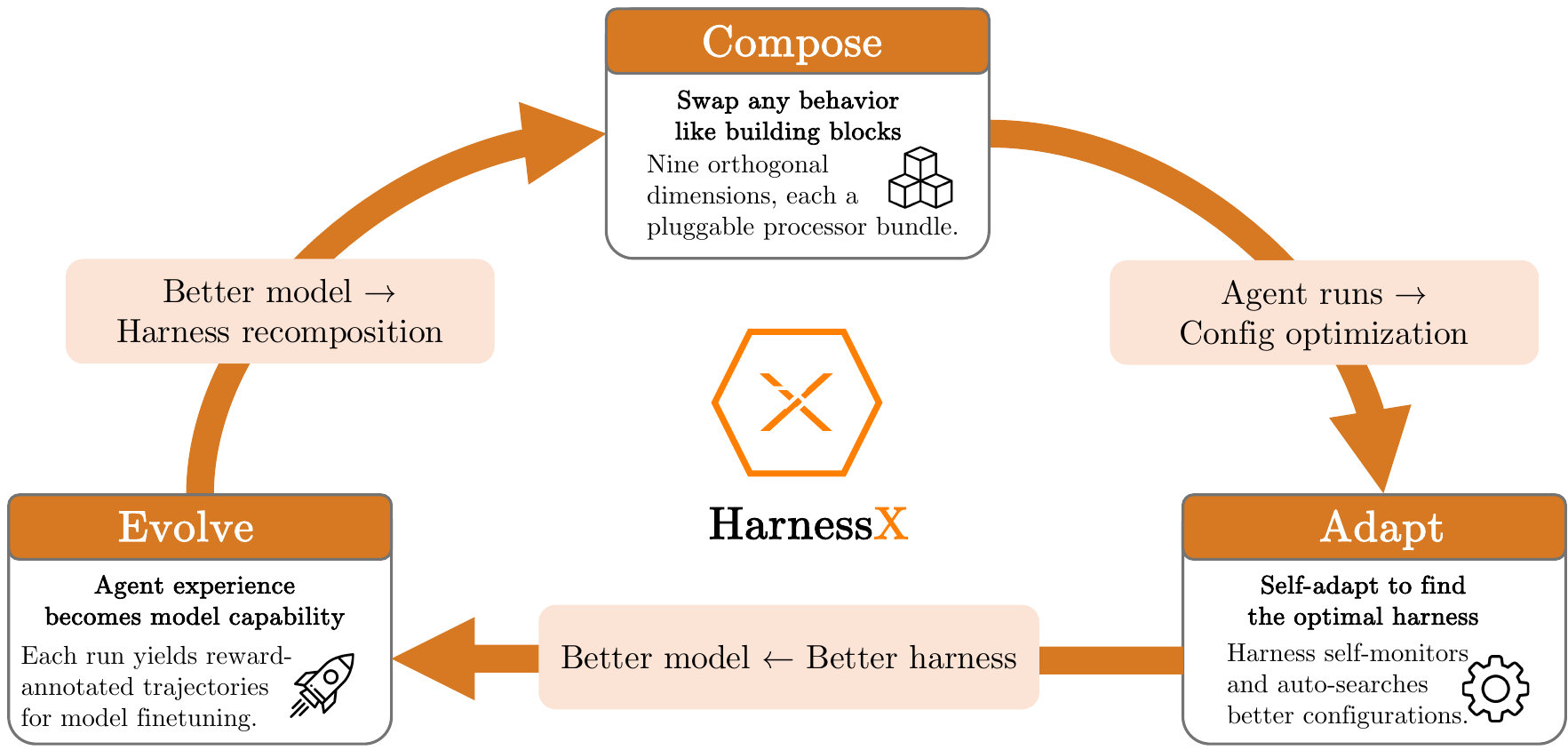}
  \captionof{figure}{\ProjectNameTitle  overview.}
  \label{fig:three-axes}
  \par
}

\tableofcontents
\newpage

\section{Introduction}
\label{sec:intro}

The capacity of modern agents depends not only on the underlying model~\cite{deepseekai2026deepseekv4,glm5team2026glm5vibecodingagentic, yang2025qwen3,team2023gemini}, but on the mediation imposed by the surrounding \textit{harness}~\cite{lu2026openclaw,liagent,claudecode}. This harness converts raw model outputs into structured agent behaviors by determining how tasks are represented, how external services are accessed, and how intermediate decisions are communicated during execution. As agents tackle longer-horizon tasks in richer environments, harness design becomes integral to agent development.

Despite this importance, harness development remains far from a mature engineering discipline. \textit{First}, harnesses are hand-engineered and static: any change in model version, tooling, or problem domain requires bespoke modification, with no mechanism for experience-driven improvement. \textit{Second}, harnesses are architecturally entangled: they typically combine prompt templates, tool wrappers, retry policies, and memory in the same codepaths, so changes to one component silently break others, and reuse across domains reduces to copying rather than composition. \textit{Third}, harness engineering and model training operate independently: trajectory data collected while improving the harness is discarded rather than incorporated into model training, and model improvements do not translate into harness improvements.

We address these gaps by treating the harness as a \textit{first-class object} that can be composed, adapted, and evolved alongside the model. \ProjectName embodies this principle as a unified harness foundry. It begins with a modular foundation: harness primitives spanning context, tools~\cite{feng2025retool}, skills, control, and memory are described via typed interfaces and composed via a substitution algebra. This separates concerns that existing systems typically conflate. On top of this substrate, we introduce AEGIS, an observability-driven and auditable harness adaptation engine. Framing harness adaptation not as ad-hoc editing but as a learning problem over symbolic artifacts (prompts~\cite{zhou2025proposer}, tools, memory, and control policies) reveals that standard RL pathologies (reward hacking, catastrophic forgetting~\cite{kirkpatrick2017overcoming}, under-exploration~\cite{ladosz2022exploration}) become concrete design risks. To address these risks, AEGIS combines full trace observability with a four-stage pipeline (Digester, Planner, Evolver, and Critic) that compresses traces, plans adaptations, generates candidates, and assesses changes. Finally, we close the loop between harness adaptation and model training via \textbf{harness-model co-evolution}. Traces produced during harness adaptation serve as reinforcement-learning signal for model training, so that model improvements feed back into subsequent harness evolution.

We empirically validate \textbf{\ProjectName} across five benchmarks (GAIA, ALFWorld, WebShop, $\tau^3$-Bench, SWE-bench Verified), three task-agent families (Claude Sonnet 4.6, GPT-5.4, Qwen3.5-9B), and up to 15 evolution rounds. Harness evolution yields an average absolute gain of +14.5\% across 15 model--benchmark configurations, with individual gains ranging from 0.0\% to +44.0\% among improving configurations (14 of 15), from +1.1\% ($\tau^3$-Bench, near-ceiling baseline) to +44.0\% (ALFWorld, weakest agent). Gains exhibit an inverse-scaling pattern: on ALFWorld and GAIA, the weakest task agent benefits most (+44.0\% for Qwen3.5-9B vs.\ +11.2\% for Sonnet 4.6 on ALFWorld), suggesting that evolved harnesses address behavioral gaps that weaker models cannot self-correct. On heterogeneous task sets (GAIA), single-harness evolution stagnates; a variant-isolation ablation restores stable improvement (+13.6\%, non-degrading over 15 rounds).

In summary, our contributions are four-fold:
\begin{itemize}
  \item \textbf{Harness Composition} (Section~\ref{sec:method-foundry}). We formalize the harness as a first-class, typed object composed of processors attached to lifecycle hooks. A nine-dimensional taxonomy spans the full behavioral space, and a substitution algebra enables per-task configuration with type-safe insertion and removal. This compositional structure makes the intended scope of each behavioral change explicit---a precondition for the variant isolation that stabilizes evolution.
  \item \textbf{Harness Adaptation} (Section~\ref{sec:method-aegis}). We introduce AEGIS, a trace-driven, multi-agent harness evolution engine. An operational mirror maps harness adaptation onto standard RL constructs, converting familiar RL pathologies (reward hacking, catastrophic forgetting, under-exploration) into concrete design risks addressed by a four-stage pipeline (Digester, Planner, Evolver, Critic) with deterministic gating. An optional variant-isolation strategy prevents cross-task interference on heterogeneous benchmarks.
  \item \textbf{Harness-Model Co-Evolution} (Section~\ref{sec:method-coevolution}). We close the optimization loop by interleaving harness evolution with model reinforcement learning over a shared replay buffer. Cross-harness GRPO enables the model to internalize strategies from successive harness versions, breaking the scaffolding ceiling that limits harness-only adaptation and the training-signal ceiling that limits model-only RL.
  \item \textbf{Empirical Validation} (Section~\ref{sec:experiments}). Across five benchmarks, three task-agent families, and up to 15 evolution rounds, \ProjectName yields an average gain of +14.5\% (up to +44.0\%), with gains largest where baselines are lowest. A variant-isolation ablation resolves stagnation on heterogeneous task sets, and co-evolution yields an additional +4.7\% over harness-only evolution (Section~\ref{sec:exp-coevolution}).
\end{itemize}
\section{Related Work}
\label{sec:related}

\subsection{Harness Engineering}
\label{sec:related-harness-eng}

Existing agent infrastructure occupies a spectrum of increasingly opinionated harness abstractions. At the primitive layer, libraries such as \texttt{LangChain}~\cite{langchain}, \texttt{LlamaIndex}~\cite{Liu_LlamaIndex_2022}, and \texttt{Smolagents}~\cite{smolagents} provide typed building blocks for prompts, tools, retrieval, and memory. These primitives can be tested in isolation but do not support harness-level composition: two harnesses built from identical primitives may still differ in structure.

The next level of abstraction orchestrates these primitives into reusable patterns. \texttt{LangGraph}~\cite{langgraph} models the behavior of an agent with a stateful graph; \texttt{AutoGen}~\cite{wu2024autogen} models multi-agent interaction as structured conversation; \texttt{CrewAI}~\cite{moura2025crewai} assigns role-based identities to agents; and \texttt{Letta}~\cite{packer2023memgpt} couples autonomous loops with persistent memory. Although these frameworks make harness writing easier, they impose a particular control loop, so combining patterns, replacing components, and porting enhancements across tasks mostly remain manual.

Lastly, there are productized, domain-specific harnesses such as \texttt{Claude Code}~\cite{claudecode}, \texttt{Cursor}~\cite{cursor}, \texttt{Manus}~\cite{shen2025mind}, and \texttt{DeerFlow}~\cite{deerflow}. These systems demonstrate the impact of harness design but remain architecturally static, evolving only through manual iteration.

Two structural gaps persist across all three layers. First, no layer exposes the harness as a substitutable entity composed of typed elements, so building a per-task harness always involves rewriting. Second, no mechanism exists for in-loop improvement: once defined, a harness evolves only through human iteration between releases.

Concurrently, Claude Code introduced \textit{Dynamic Workflows}~\cite{anthropic2026dynamicworkflows}, enabling the model to generate task-specific harness scripts at runtime. While this represents a step toward adaptive harnesses, it operates within a single session without persistent trace-based optimization, cross-session evolution, or harness--model co-training. \ProjectName addresses both gaps by treating harness adaptation as a multi-round, trace-driven learning problem with typed composition for variant isolation, structured observability for pathology detection, and a shared replay buffer that closes the loop between harness evolution and model training.

\subsection{Self-Evolving Agents}
\label{sec:related-self-evolving}

Research on self-evolving agents investigates how an agent system can improve without retraining the underlying foundation model. Early work focused on the single most easily editable aspect: the prompt. Approaches like \texttt{APE}~\cite{zhou2022large}, \texttt{OPRO}~\cite{yang2024large}, \texttt{EvoPrompt}~\cite{guo2024connecting}, \texttt{Promptbreeder}~\cite{fernando2024promptbreeder} treat instruction formulation as a black-box optimization problem, while \texttt{ProTeGi}~\cite{pryzant2023automatic} and \texttt{TextGrad}~\cite{yuksekgonul2024textgrad} introduce gradient-inspired textual feedback to make the optimization process explicit. \texttt{DSPy}~\cite{khattab2023dspy} and \texttt{MIPRO}~\cite{opsahl2024optimizing} extend this approach by compiling a declarative LM program, whose prompts are optimized against labeled data. These approaches establish instructions as a learnable component, but harness-level features (tools, memory, control flow) remain outside the optimization scope.

Another line of work improves agents by accumulating and reusing prior execution experience in memory: \texttt{Memento}~\cite{zhou2025memento} improves agents through case-based memory without fine-tuning the model, while \texttt{MIA}~\cite{qiao2026memory} unifies non-parametric and parametric memory within a single Manager-Planner-Executor framework: a non-parametric store of compressed trajectories and a parametric planner that evolves on the fly at test time, coupled by a bidirectional loop that continually converts experience between the two, demonstrating superiority across eleven benchmarks.

Subsequent works extend optimization to agent workflows. \texttt{GPTSwarm}~\cite{zhuge2024gptswarm}, \texttt{ADAS}~\cite{hu2025automated}, \texttt{AFlow}~\cite{zhang2025aflow}, \texttt{A$^2$Flow}~\cite{zhao2026a2flow}, \texttt{AgentSwift}~\cite{li2026agentswift}, \texttt{ResMAS}~\cite{zhou2026resmas}, and \texttt{EvoAgentX}~\cite{wang2025evoagentx} search over collaboration strategies, agent ordering, and aggregation mechanisms. These works demonstrate that workflow structure is learnable and yields larger gains than prompt-only optimization. However, component-level artifacts (tool implementations, memory policies, node-internal prompts) remain static: the optimization scope covers inter-component relations but does not encompass the full harness.

A final group treats harness evolution explicitly. \texttt{SICA}~\cite{robeyns2025self} optimizes a SWE-bench agent's source code directly, while \texttt{Darwin Gödel Machine}~\cite{lange2025darwin} proposes open-ended optimization over a database of agent variants. \texttt{HyperAgents}~\cite{zhang2026hyperagents} makes the optimization process itself adaptable; \texttt{Meta-Harness}~\cite{lee2026meta} improves sampling efficiency via a file-system-based interface. \texttt{AHE}~\cite{lin2026agentic} and \texttt{Life-Harness}~\cite{xu2026adapting} emphasize observability, explainability, and source-code rewriting. Collectively, these works establish the harness as an evolutionary target and demonstrate that observability is essential for stable self-improvement. However, their designs lack a unifying theoretical framework that connects observed failure modes to principled defenses.

The heuristic-learning theory~\cite{weng2026learning_beyond_gradients} partially addresses this gap by mapping RL concepts to symbolic self-optimization updates. In this framework, observable traces correspond to proper credit assignment, falsifiable change manifests correspond to reward shaping, and proposal-critique cycles provide structured exploration. \ProjectName instantiates this paradigm, formalizing the correspondence as the \textit{operational mirror} between RL and symbolic harness evolution (Section~\ref{sec:aegis-mirror}).
\section{Harness Composition}
\label{sec:method-foundry}

The gap identified in Section~\ref{sec:related-harness-eng} is the absence of an infrastructure layer that exposes the harness as a typed, substitutable entity. Primitive libraries leave composition to application code, orchestrators expose a fixed set of patterns, and product harnesses are opaque end-to-end. Without a compositional substrate, every behavioral change or cross-team handoff requires re-implementation. \ProjectName addresses this via a unified design principle: the harness is a first-class value, the processor is a typed atomic component, and composition proceeds via processor insertion at typed hook points. We formalize the harness (Section~\ref{sec:foundry-object}), its building block, the processor (Section~\ref{sec:foundry-processor}), and the nine-dimensional processor taxonomy (Section~\ref{sec:foundry-taxonomy}). Definitions are intentionally concise: their role is to establish the vocabulary and expose the edit surface on which harness evolution (Section~\ref{sec:method-aegis}) operates.

\begin{table}[!htbp]
\centering
\small
\begin{tabular}{lll}
\toprule
Hook & Event type & Permitted modifications \\
\midrule
\texttt{task\_start}    & \texttt{TaskStartEvent}     & system prompt \\
\texttt{step\_start}    & \texttt{StepStartEvent}     & structural history edits \\
\texttt{before\_model}  & \texttt{BeforeModelEvent}   & last user content; one user-message append \\
\texttt{after\_model}   & \texttt{ModelResponseEvent} & response content, tool calls \\
\texttt{before\_tool}   & \texttt{ToolCallEvent}      & tool input, approval flag \\
\texttt{after\_tool}    & \texttt{ToolResultEvent}    & tool result \\
\texttt{step\_end}      & \texttt{StepEndEvent}       & read-only \\
\texttt{task\_end}      & \texttt{TaskEndEvent}       & read-only \\
\bottomrule
\end{tabular}
\caption{Hook points and their permitted modifications.}
\label{tab:hook-points}
\end{table}

\subsection{The Harness as a First-Class Object}
\label{sec:foundry-object}

A harness in \ProjectName is the pair $\mathcal{H} = (\mathcal{M}, \mathcal{C})$, where $\mathcal{M}$ is a model configuration and $\mathcal{C}$ is a harness configuration. The two address disjoint concerns: $\mathcal{M}$ records \textit{which} model serves which role (\texttt{main}, \texttt{judge}, \texttt{evaluator}) and the fallback policy for each role; $\mathcal{C}$ records \textit{how} the agent behaves independently of model identity. They combine into an executable agent via \texttt{agent = model\_config.agentic(harness\_config)}: an agent in \ProjectName is a processor pipeline bound to a model, both independently substitutable.

The harness configuration itself decomposes as $\mathcal{C} = (\mathbf{P}, \mathbf{S})$. $\mathbf{P} : \mathcal{H}\!\mathit{ook} \to \mathrm{List}[\mathit{Processor}]$ is a hook-indexed list of processors, where $\mathcal{H}\!\mathit{ook}$ is the eight-element set of lifecycle events in Table~\ref{tab:hook-points}. $\mathbf{S}$ is a fixed set of orthogonal \textit{slot resources}: tool registry, tracer, workspace, sandbox provider, and plugin list. Slots are singletons, shared across all processors in a configuration; processor state is instance-private. $\mathbf{P}$ implements all per-step behavior; $\mathbf{S}$ houses the shared infrastructure that processors depend on but do not own.

We call $\mathcal{C}$ a first-class object because it is independently serializable, comparable, hashable, and substitutable. Two agents sharing $\mathcal{C}$ but differing in $\mathcal{M}$ execute the same processor pipeline, with behavior differing only in model responses; two agents sharing $\mathcal{M}$ but differing in $\mathcal{C}$ are behaviorally distinct. This reification is the precondition for programmatic evolution (Section~\ref{sec:method-aegis}).

\subsection{The Processor Abstraction}
\label{sec:foundry-processor}

Every per-step behavior in \ProjectName is implemented as a \textit{processor}, an object satisfying the protocol \texttt{async def process(self, event: Event) -> AsyncIterator[Event]}. A processor consumes one event and yields zero or more, producing exactly one of five outcomes: pass-through (yield unchanged), transform (yield modified), split (yield multiple same-type events, processed independently downstream), intercept (yield nothing, blocking propagation), or interrupt (raise an exception, which halts the loop). This restricted interface enables compositionality: every processor at a given hook consumes and yields the same event type, so processors compose by sequential application and can be inserted or removed without affecting type correctness of the surrounding pipeline.

As listed in Table~\ref{tab:hook-points}, processors attach to one of eight hook points emitted by the run loop. The run loop validates hook contracts after each invocation: a violation (e.g., modifying a read-only field) raises an exception immediately rather than silently propagating corrupted state. Each processor carries three class-level metadata fields that govern composition: \texttt{\_singleton\_group} names a mutual-exclusion class, ensuring at most one processor per group; \texttt{\_order} is an ordering hint within a hook (with constants \texttt{PRE}, \texttt{NORMAL}, \texttt{POST}); and \texttt{\_after} is a list of soft dependencies on other singleton groups.

This design makes harness evolution a first-class operation: AEGIS can insert a new processor at a specific hook, replace an existing one by matching its singleton group, or remove a processor entirely---all without touching other processors at the same or different hooks. Because the type contract (input event type $=$ output event type) is enforced per-hook, any such substitution preserves the well-typedness of the overall pipeline. The metadata fields further constrain composition: \texttt{\_singleton\_group} prevents conflicting duplicates, and \texttt{\_order} ensures that newly inserted processors interact predictably with existing ones. These guarantees are the mechanism by which variant isolation (Section~\ref{sec:aegis-ensemble}) operates---each variant differs only in which processors occupy which hooks, and the type system ensures that no variant can silently violate the pipeline contract during evolution.

\subsection{The Nine-Dimensional Taxonomy}
\label{sec:foundry-taxonomy}

We organize the behavioral space along nine dimensions: \textit{model selection} (D1) decides which model serves which role; \textit{context assembly} (D2) determines what is presented to the model at each step; \textit{memory management} (D3) governs what carries across steps and sessions; \textit{tool ecosystem} (D4) controls which tools the agent can invoke; \textit{execution environment} (D5) determines where tool-induced side-effects materialize; \textit{evaluation and reward} (D6) specifies how outcomes are judged; \textit{control and safety} (D7) enforces rules that keep the agent from looping, overspending, or drifting from intent; \textit{observability} (D8) records each event, model call, and tool invocation; and the \textit{training bridge} (D9) converts execution trajectories into reinforcement-learning records. Figure~\ref{fig:aegis-loop} illustrates the full taxonomy along with representative processors and the hooks at which they typically attach in a standard configuration.

In practice, AEGIS edits span all nine dimensions during evolution: D2 (context assembly) and D4 (tool ecosystem) are the most frequent edit targets (Section~\ref{sec:exp-main}), while D8 (observability) provides the trace substrate on which AEGIS itself reasons, and D9 (training bridge) supplies trajectory records for co-evolution (Section~\ref{sec:method-coevolution}), closing the optimization loop.


\begin{table}[!htbp]
\centering
\small
\begin{tabular}{lll}
\toprule
\textbf{RL concept} & \textbf{Symbolic-space dual} & \textbf{AEGIS realization} \\
\midrule
Policy $\pi$ & Harness-update procedure $\pi_{\mathrm{evo}}$ & Four-stage pipeline (Section~\ref{sec:aegis-arch}) \\
State $s_t$ & $(\mathcal{H}_t, \mathcal{T}_t)$ & Harness configuration + trace store \\
Action $a_t$ & Typed harness edit & Builder operation + change manifest \\
Feedback & Trace $\tau$ + verifier score $r$ & Observability layer \\
Update & $\mathcal{H}_{t+1} \leftarrow U(\widetilde{\mathcal{H}}_{t}, \mathcal{T}_t, r_t)$ & Deterministic acceptance gate \\
\bottomrule
\end{tabular}
\caption{Operational mirror: RL concepts and their symbolic-space duals in AEGIS.}
\label{tab:operational-mirror}
\end{table}

\section{Harness Adaptation}
\label{sec:method-aegis}

\begin{figure}[!htbp]
  \centering
  \includegraphics[width=\columnwidth]{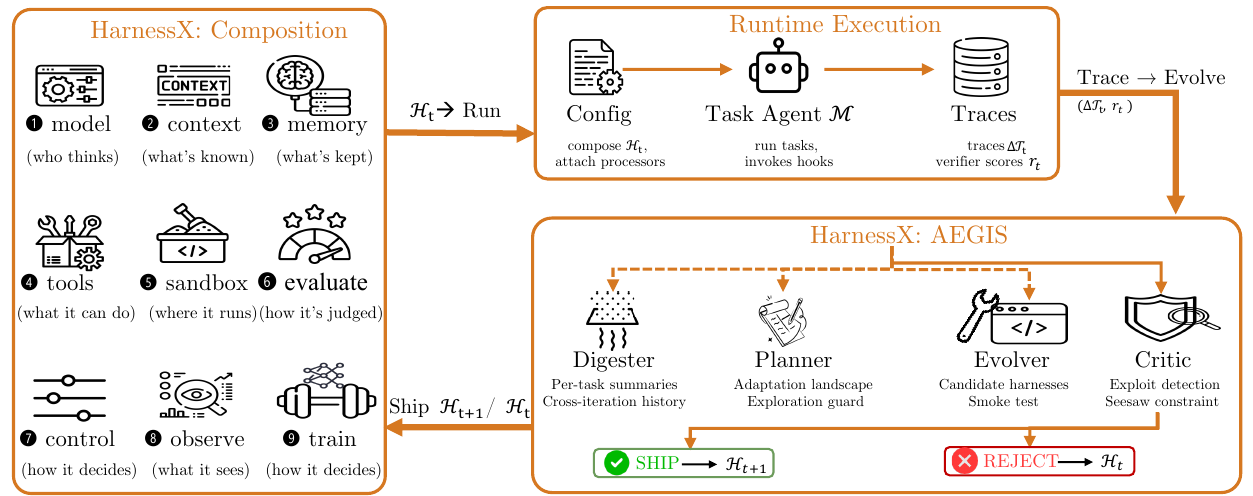}
  \caption{The AEGIS evolution loop. A single meta-agent $\mathcal{M}$ drives all four stages (Digester, Planner, Evolver, Critic), selectively invoking each based on whether sufficient signal exists to continue. A deterministic gate ships or rejects the candidate edit.}
  \label{fig:aegis-loop}
\end{figure}

The composition layer (Section~\ref{sec:method-foundry}) provides a typed, substitutable harness; as illustrated in Figure~\ref{fig:aegis-loop}, AEGIS is the system that evolves it. The key insight is that harness evolution maps structurally onto reinforcement learning in a symbolic space: harness configurations are states, typed edits are actions, and execution traces plus verifier scores constitute feedback. This mapping is predictive: it identifies three failure modes analogous to known RL pathologies (reward hacking, catastrophic forgetting, under-exploration) that motivate AEGIS's architectural defenses and are empirically confirmed in Section~\ref{sec:exp-failure}.

We formalize the correspondence (Section~\ref{sec:aegis-mirror}), analyze the pathologies it predicts (Section~\ref{sec:aegis-pathologies}), derive the four-stage pipeline as a defense architecture (Section~\ref{sec:aegis-arch}), present the adaptation loop (Section~\ref{sec:aegis-loop}), and introduce variant isolation for stable multi-variant evolution (Section~\ref{sec:aegis-ensemble}).

\subsection{The Operational Mirror}
\label{sec:aegis-mirror}

We formalize harness evolution as an MDP over symbolic artifacts. Table~\ref{tab:operational-mirror} summarizes the mapping; we first state three definitions that ground the correspondence.

\paragraph{Definition 1 (Harness Configuration).}
A harness configuration is a tuple $\mathcal{H} = (c_1, c_2, \ldots, c_9)$, where each $c_i \in \mathcal{C}_i$ instantiates one of the nine behavioral dimensions (Section~\ref{sec:foundry-taxonomy}): model selection ($c_1$), context assembly ($c_2$), memory management ($c_3$), tool ecosystem ($c_4$), execution environment ($c_5$), evaluation and reward ($c_6$), control and safety ($c_7$), observability ($c_8$), and training bridge ($c_9$). Each $\mathcal{C}_i$ is the set of valid processor configurations for dimension $i$, constrained by hook-type contracts and singleton-group exclusion (Section~\ref{sec:foundry-processor}).

\paragraph{Definition 2 (Harness Edit).}
A harness edit is a function $e: \mathcal{H} \to \mathcal{H}$ that modifies one or more dimensions while preserving type contracts. The action space $\mathcal{E}$ is \textit{discrete but open-ended}: each edit is a code-level artifact (new processor source, modified prompt template, reconfigured tool registry, or control-flow rewrite) generated by the meta-agent LLM, not selected from a pre-enumerated set. Combinatorial explosion is managed not by exhaustive search but by the LLM's generative capacity---the Planner proposes edits from trace-grounded hypotheses---and by type constraints that prune invalid compositions at generation time.

\paragraph{Definition 3 (Operational Mirror).}
The operational mirror is the tuple $(\mathcal{H}, \mathcal{E}, \mathcal{R}, \mathcal{T})$, where $\mathcal{H}$ is the harness-configuration space (states), $\mathcal{E}$ is the code-level edit space (actions), $\mathcal{R}: \mathcal{H} \times \mathcal{E} \to \mathbb{R}$ maps a configuration--edit pair to a scalar reward (verifier scores aggregated over an adaptation batch), and $\mathcal{T}$ is the trace store that provides structured feedback beyond the scalar signal. This tuple forms an MDP at the harness level: harness configurations are states, typed edits are actions, execution traces plus verifier scores constitute feedback, and a deterministic acceptance gate governs state transitions.

\paragraph{MDP instantiation.}
Let $\mathcal{H}_t$ denote the harness configuration at iteration $t$ (the model $\mathcal{M}$ is fixed throughout evolution), and let
$\mathcal{T}_t$ denote the trace store accumulated from all previous executions. We
define the symbolic state as $s_t=(\mathcal{H}_t,\mathcal{T}_t)$. A harness-update
policy $\pi_{\mathrm{evo}}$ selects an action
$a_t \sim \pi_{\mathrm{evo}}(\cdot \mid s_t)$, where $a_t \in \mathcal{E}$ is a code-level edit drawn
from the builder algebra. Applying this edit yields a candidate harness
$\widetilde{\mathcal{H}}_{t}=a_t(\mathcal{H}_t)$. Running the candidate on an adaptation batch (with the fixed model $\mathcal{M}$) produces new traces $\Delta\mathcal{T}_t$ and per-task verifier scores $r_t$. A deterministic acceptance operator
$U(\widetilde{\mathcal{H}}_{t}, \mathcal{T}_t, r_t)$ then either commits the candidate
($\mathcal{H}_{t+1}=\widetilde{\mathcal{H}}_{t}$) or rejects it
($\mathcal{H}_{t+1}=\mathcal{H}_t$), enforcing the seesaw constraint: the candidate must not regress any previously solved task recorded in $\mathcal{T}_t$. In both cases, the trace store grows:
$\mathcal{T}_{t+1}=\mathcal{T}_t \cup \Delta\mathcal{T}_t$.

This MDP operates at the harness level: within a single task, $\mathcal{H}_t$ (together with the fixed $\mathcal{M}$) determines the agent's behavior; across iterations, the harness-update policy $\pi_{\mathrm{evo}}$ modifies the harness. AEGIS realizes $\pi_{\mathrm{evo}}$ as a four-stage pipeline (Digester, Planner, Evolver, Critic) that maps $s_t$ to candidate edits through trace compression, adaptation planning, edit generation, and candidate assessment.

\subsection{Pathologies in Symbolic Space}
\label{sec:aegis-pathologies}

The mirror is not merely an analogy; it converts reinforcement-learning concepts into design requirements. We refer to three well-documented failure modes in RL, namely reward hacking~\cite{guo2025deepseek}, catastrophic forgetting~\cite{kirkpatrick2017overcoming}, and under-exploration~\cite{ladosz2022exploration}, collectively as \textit{RL pathologies}. Once harness adaptation is cast as an MDP over symbolic artifacts, these pathologies reappear in amplified form, shaped by two properties of the symbolic setting: (1)~a language-model evolver can construct structured exploits that numerical parameter perturbations cannot express, and (2)~edits to shared components propagate non-locally through the harness. Each pathology below motivates a corresponding architectural defense in Section~\ref{sec:aegis-arch}.

\paragraph{Reward hacking.}
In standard RL, reward hacking~\cite{guo2025deepseek} exploits loopholes in the reward signal without genuine task completion. Symbolic harness evolution amplifies this risk because the evolver can target the verification protocol directly: embedding benchmark answers into prompts, exploiting format regularities in the verifier, or introducing a processor that rewrites outputs to match verifier expectations.

\paragraph{Catastrophic forgetting.}
Catastrophic forgetting~\cite{kirkpatrick2017overcoming} occurs when improving performance on one region of the task distribution harms another. In symbolic harness evolution, an edit that repairs failure pattern $A$ can silently regress pattern $B$, because effects propagate through shared context, tools, memory policies, and control rules. Without explicit regression checking, an evolver conditioned only on failing-task traces cannot distinguish local gain from global regression.

\paragraph{Under-exploration.} Under-exploration~\cite{ladosz2022exploration} manifests as a bias toward low-risk local edits: prompt rephrasing, tool-description tuning, or minor control-flow tweaks. These edits are cheap to generate and frequently pass gating without regressing solved tasks, biasing subsequent Planner hypotheses toward the same edit neighborhood. Structural changes (decomposing one agent into several, replacing the control strategy, or adopting a new memory architecture) require deliberate hypothesis formation and rarely emerge from trace-conditional local repair. Without a mechanism to propose edits beyond the immediate failure neighborhood, the system plateaus once local edits are exhausted.

\paragraph{Summary.} Symbolic harness evolution inherits the structural risks of RL (reward hacking, catastrophic forgetting, and under-exploration), and AEGIS addresses each with a dedicated mechanism: the Critic (reward hacking), the deterministic gating layer (catastrophic forgetting), and the Planner (under-exploration).

\begin{algorithm}[H]
\small
\caption{AEGIS Harness Evolution Loop (selective invocation)}
\label{alg:aegis-loop}
\SetAlgoLined
\LinesNumbered
\SetKwInput{KwInput}{Input}
\SetKwInput{KwOutput}{Output}
\KwInput{Initial harness $\mathcal{H}_0$, meta-agent $\mathcal{M}$, budget $T$, patience $P$, threshold $\alpha$}
\KwOutput{Evolved harness $\mathcal{H}_{t+1}$, trace store $\mathcal{T}_{t+1}$}
$\mathcal{T}_0 \leftarrow \emptyset$\; $\mathit{idle} \leftarrow 0$\;
\For{$t = 0, 1, \ldots, T{-}1$}{
  Sample batch $B_t$\; run $\mathcal{H}_t$ on $B_t$ to get traces $\Delta\mathcal{T}_t$\; $\mathcal{T}_{t+1} \leftarrow \mathcal{T}_t \cup \Delta\mathcal{T}_t$\;
  \tcc{Digester (selective)}
  $(\mathit{evidence}_t,\; a_t) \leftarrow \mathcal{M}.\textsc{Digester}(\Delta\mathcal{T}_t,\; \mathcal{T}_t)$\;
  \lIf{$a_t < \alpha$}{$\mathcal{H}_{t+1} \leftarrow \mathcal{H}_t$\; $\mathit{idle}{+}{+}$\; \textbf{continue}}
  \tcc{Planner (selective)}
  $\mathit{landscape}_t \leftarrow \mathcal{M}.\textsc{Planner}(\mathit{evidence}_t)$\;
  \lIf{$\mathit{landscape}_t = \emptyset$}{$\mathcal{H}_{t+1} \leftarrow \mathcal{H}_t$\; $\mathit{idle}{+}{+}$\; \textbf{continue}}
  \tcc{Evolver (selective)}
  $\{(\widetilde{\mathcal{H}}_{t}^{\,k},\, \mathrm{manifest}_k)\}_{k=1}^{K_t} \leftarrow \mathcal{M}.\textsc{Evolver}(\mathcal{H}_t,\; \mathit{landscape}_t)$\;
  \lIf{$K_t = 0$}{$\mathcal{H}_{t+1} \leftarrow \mathcal{H}_t$\; $\mathit{idle}{+}{+}$\; \textbf{continue}}
  \tcc{Critic \& Gate (mandatory)}
  $\mathit{ranking} \leftarrow \mathcal{M}.\textsc{Critic}(\{(\widetilde{\mathcal{H}}_{t}^{\,k},\, \mathrm{manifest}_k)\},\; \mathit{evidence}_t)$\;
  $k^\star \leftarrow \bot$\;
  \ForEach{$k$ \textbf{in} $\mathit{ranking}$}{
    \lIf{\textnormal{DeterministicGate}$(\widetilde{\mathcal{H}}_{t}^{\,k},\, \mathcal{H}_t,\, \mathcal{T}_t)$ passes}{$k^\star \leftarrow k$\; \textbf{break}}
  }
  \lIf{$k^\star \neq \bot$}{$\mathcal{H}_{t+1} \leftarrow \widetilde{\mathcal{H}}_{t}^{\,k^\star}$\; $\mathit{idle} \leftarrow 0$}
  \lElse{$\mathcal{H}_{t+1} \leftarrow \mathcal{H}_t$\; $\mathit{idle}{+}{+}$}
  \lIf{$\mathit{idle} \geq P$}{\textbf{break}}
}
\Return $\mathcal{H}_{t+1},\; \mathcal{T}_{t+1}$
\end{algorithm}

\subsection{AEGIS Architecture}
\label{sec:aegis-arch}

AEGIS is the harness-evolution engine of \ProjectName. It comprises four stages arranged in a predefined workflow---Digester, Planner, Evolver, Critic---all driven by the same meta-agent LLM, which \textit{selectively invokes} them: no external router decides stage execution; instead, the meta-agent itself determines at each stage whether sufficient signal exists to continue. The Digester, Planner, and Evolver each evaluate a continuation condition and may short-circuit the round (below-threshold actionability, empty landscape, or zero viable candidates), while the Critic together with the deterministic gating layer is mandatory for every candidate that reaches it. No edit can ship without passing through the Critic and gate. The division of labor across stages addresses the pathologies of Section~\ref{sec:aegis-pathologies}: the Digester compresses raw traces into structured task-level evidence; the Planner constructs an adaptation landscape spanning both incremental and structural changes; the Evolver produces typed builder edits with explicit change manifests; and the Critic, together with the gating layer, rejects edits whose claimed improvement lacks trace support or whose acceptance would regress previously solved tasks.

All stages share a single information substrate: the trace store, a structured record of execution events, verifier-scored outcomes, regression signals, and shipped or rejected edits. No stage consumes input beyond the trace store and the current harness $\mathcal{H}_t$. Data flows forward through the pipeline with selective gating: the Digester may determine that no actionable failures exist (all tasks pass or signal is too sparse), terminating the round immediately; the Planner may find no viable adaptation landscape given the current evidence and edit history; and the Evolver may produce no type-safe candidates. In each case the round exits cleanly with a no-op outcome. Only the Critic and deterministic gate are unconditional: any candidate that survives the upstream stages must pass through both before shipping. The Critic may additionally issue a single revision request to the Evolver before returning its final verdict.

\paragraph{Digester.}
A single iteration on GAIA (103 tasks, pass@2) generates ${\sim}$10M tokens of raw traces: model reasoning steps, tool invocations with their outputs, and timing metadata. Passing this volume directly to downstream stages exceeds context limits, yet naive truncation discards diagnostic signal. The Digester compresses each task's traces into a structured per-task summary: binary outcome, failure category (if any), implicated component identifiers, and supporting evidence excerpts. It also provides cross-iteration continuity: each task's summary links to its history of prior outcomes and shipped edits, enabling the Planner to distinguish persistent failures from transient noise.

\paragraph{Planner.}
The Planner receives the Digester's output (task-level summaries enriched with cross-iteration history) and constructs an adaptation landscape: which tasks are failing, what edits have been attempted, which components are implicated, and which edit types (prompt, tool, processor, configuration) remain untried. This stage is the primary defense against under-exploration: by constructing the landscape before edit generation, it prevents the pipeline from converging on trace-conditional local repair, ensuring that structural changes (tool additions, processor rewrites, memory-policy redesigns) are considered alongside incremental prompt edits.

\paragraph{Evolver.}
Given the Planner's adaptation landscape, the Evolver produces one or more candidate harnesses $\{\widetilde{\mathcal{H}}_{t}^{\,k}\}_{k=1}^{K_t}$, each specified as a typed builder operation on the current harness $\mathcal{H}_t$. Each candidate carries a \textit{change manifest}: the edited components, the intended behavioral effect, and the tasks expected to improve or regress. When introducing new processor code, the Evolver must also provide a smoke test confirming that the processor instantiates and runs on synthetic input without raising exceptions. The builder algebra guarantees type-safety (every candidate satisfies hook-type contracts and processor-composition rules) but not behavioral safety; an edit that type-checks may still produce non-local behavioral effects, detectable only by the Critic and gating layer.

\paragraph{Critic and gating.}
The Critic defends against reward hacking; the deterministic gating layer defends against catastrophic forgetting. The Critic evaluates each candidate by comparing its change manifest against trace evidence and assessing whether edits risk non-local effects through shared state or control flow. When gaps are detected, it issues a single revision request to the Evolver. After at most one revision cycle, the Critic returns either \texttt{no\_op} or an ordered \texttt{ship\_ranking}. The deterministic gate then applies acceptance checks in sequence: manifest completeness, configuration normalization (ensuring the candidate is in canonical form), build or smoke tests (when applicable), and the seesaw constraint (regression check on previously passing tasks; Section~\ref{sec:aegis-mirror}). The first failing check halts the sequence; passing candidates are committed and failing ones archived with their rejection reason. This decouples LLM judgment from acceptance: regardless of the Critic's recommendation, only deterministic checks govern shipping.

\paragraph{Design principle.}
Language-model subagents explore, hypothesize, and propose; typed structure and deterministic gates determine what ships. This separation ensures that safety properties (no regression, no unaudited edits) hold regardless of LLM subagent failure modes.

\subsection{The Adaptation Loop}
\label{sec:aegis-loop}

Algorithm~\ref{alg:aegis-loop} formalizes the adaptation loop (each iteration corresponds to one ``round'' in Section~\ref{sec:experiments}). Starting from an initial harness $\mathcal{H}_0$, each iteration executes the current harness on an adaptation batch and selectively invokes the four stages: the Digester, Planner, and Evolver each gate on a continuation condition (sufficient actionability, non-empty landscape, and at least one type-safe candidate, respectively), while the Critic and deterministic gate are mandatory for any candidate that reaches them. A round commits a new harness only when a candidate clears all acceptance checks.

\subsection{Variant Isolation via Ensemble Routing}
\label{sec:aegis-ensemble}

The adaptation loop (Section~\ref{sec:aegis-loop}) maintains a single harness $\mathcal{H}_t$. When tasks require conflicting behaviors, an edit that improves one subset may regress another; the seesaw constraint rejects it, protecting stability but discarding a locally beneficial change. \textit{Variant isolation} lifts this limitation by maintaining up to $K$ harness variants $\{\mathcal{H}_t^{(1)}, \ldots, \mathcal{H}_t^{(V_t)}\}$ ($V_t \leq K$) and routing each task to the variant with the highest estimated success rate on that task's cluster across prior rounds. We term this mechanism \textit{Ensemble routing}.

The gating layer distinguishes two outcomes per candidate: (1)~the edit improves some tasks without regressing any, in which case it is applied to its target variant; or (2)~it improves a subset while regressing others, in which case the system forks a new variant rather than rejecting the edit outright (retiring the lowest-performing variant if the pool is full). Once multiple variants exist, the seesaw constraint is scoped per-variant: a candidate targeting variant $k$ is tested only against tasks routed to $k$, so improvements to one cluster cannot regress another. This design predicts three properties validated in Section~\ref{sec:exp-strategy}: (1)~non-degrading aggregate trajectory (peak = final), (2)~sustained exploration across more rounds, and (3)~lower total token consumption.

\section{Harness-Model Co-Evolution}
\label{sec:method-coevolution}

Sections~\ref{sec:method-foundry} and~\ref{sec:method-aegis} show that evolving the harness alone, with the foundation model held fixed, already delivers substantial gains, and that these gains are largest for weaker, smaller task agents, whose behavioral gaps a better harness most readily closes. Co-evolution does not displace that route; it extends it along a second axis. For a capability-limited small model, harness evolution eventually meets a \textit{scaffolding ceiling}: once the harness exposes the right tools, context, and control flow, the binding constraint becomes whether the frozen model can actually exploit them, and no harness edit can supply reasoning capacity the model itself lacks.

Symmetrically, training the model under a fixed harness meets a \textit{training-signal ceiling}: newly acquired capabilities go unexercised when the scaffold never surfaces the context, tools, or control flow that elicit them. The model is the agent's \textit{cognitive core}, supplying reasoning and planning, while the harness is its \textit{executive apparatus}, determining what the model perceives, what it can invoke, and what constrains its execution. A sharper apparatus cannot compensate for a weak core, nor a stronger core for an apparatus that never calls on it. Co-evolution targets precisely this bottleneck: by training the model within the same loop that evolves its harness, the agent improves along both axes simultaneously, breaking the ceiling that either improvement alone would leave in place. The principle of jointly evolving complementary capability components also appears in other settings: K\textsuperscript{2} Agent~\cite{wu2026k} co-evolves know-what (declarative knowledge) and know-how (procedural skill) for hierarchical mobile device control.

\begin{figure*}[!htbp]
  \centering
  \includegraphics[width=\textwidth]{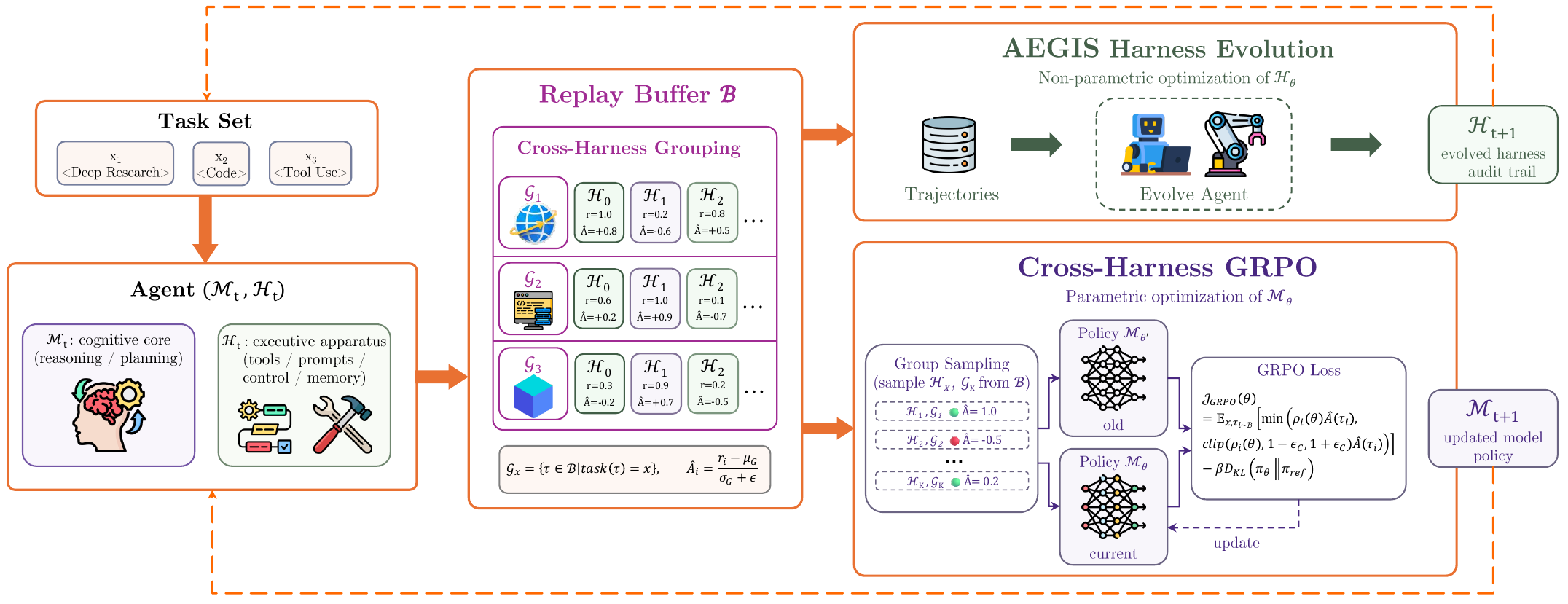}
  \caption{\textbf{The harness-model co-evolution loop.} The agent $(\mathcal{M}_t,\mathcal{H}_t)$ runs the task batch $B_t$ under a fixed verifier and the observability layer; the resulting traces and rewards $(\tau,r)$ enter a shared replay buffer $\mathcal{B}$, where cross-harness grouping pools trajectories of the same task across harness versions and computes group-relative advantages $\hat{A}$. The same buffer drives two updates over identical data: AEGIS harness evolution (Digester $\to$ Planner $\to$ Evolver $\to$ Critic, yielding the evolved harness $\mathcal{H}_{t+1}$) and \textbf{cross-harness GRPO} (group sampling and a clipped GRPO objective, yielding the updated model $\mathcal{M}_{t+1}$); both feed the next iteration.}
  \label{fig:coevolution-loop}
\end{figure*}

Figure~\ref{fig:coevolution-loop} illustrates the co-evolution mechanism. Rather than alternating between independent harness-evolution and model-training phases, \ProjectName runs both within a single iteration over a shared replay buffer. We formalize the iteration (Section~\ref{sec:coevolution-loop}), describe the two optimization substrates (Section~\ref{sec:coevolution-dual-grpo}), specify the model training objective via cross-harness GRPO (Section~\ref{sec:coevolution-grpo}), and characterize off-policy training over the shared buffer, the property that lets model RL run at no additional rollout cost (Section~\ref{sec:coevolution-offpolicy}).

\subsection{The Co-evolution Iteration}
\label{sec:coevolution-loop}

Co-evolution operates over the pair $(\mathcal{M}_t, \mathcal{H}_t)$, where $\mathcal{M}_t$ denotes trainable model parameters (relaxing the frozen-model assumption of Section~\ref{sec:method-aegis}) and $\mathcal{H}_t$ denotes the harness configuration at iteration $t$. The system maintains a fixed-capacity replay buffer $\mathcal{B}$ with first-in-first-out eviction. Each iteration proceeds as:

\begin{enumerate}[leftmargin=*]
  \item \textbf{Rollout.} Run $(\mathcal{M}_t, \mathcal{H}_t)$ on the adaptation batch $B_t$; the observability layer records each episode as a complete trace $\tau_i$, capturing every model turn, tool call, and tool result.

  \item \textbf{Verification.} A fixed verifier scores each trace into a scalar reward $r_i$. Holding the verifier fixed keeps rewards comparable across harness versions, which the cross-harness advantage (Eq.~\ref{eq:grpo-advantage}) requires.

  \item \textbf{Buffer insertion.} Append each scored trace to the shared buffer $\mathcal{B}$ together with the harness version that produced it, so successive rounds accumulate rather than overwrite; FIFO eviction keeps $\mathcal{B}$ restricted to recent rounds.

  \item \textbf{Harness evolution} ($\mathcal{H}_{t+1} \leftarrow \text{AEGIS}(\mathcal{H}_t, \mathcal{B})$, non-parametric, Section~\ref{sec:method-aegis}). The meta-agent reads the buffered traces as evidence of where the scaffold fails, proposes one discrete structural edit, and admits it only if the Critic and gating layer validate it.

  \item \textbf{Behavior log-probabilities.} For the traces just added this round, run a forward pass under the generating model $\mathcal{M}_t$ to obtain the token-level log-probabilities $\pi_{\theta_{\text{old}}}(\tau_i)$ and cache them for use in the GRPO loss; trajectories from earlier rounds reuse the values cached at their own insertion (Section~\ref{sec:coevolution-offpolicy}).

  \item \textbf{GRPO update} ($\mathcal{M}_{t+1} \leftarrow \text{GRPO}(\mathcal{M}_t, \mathcal{B})$, parametric, Section~\ref{sec:coevolution-grpo}). Partition traces into per-task groups spanning harness versions, assign each a group-relative advantage, and take a clipped policy-gradient step with a KL anchor to the fixed reference.

  \item \textbf{Advance.} Return to step~1 with the evolved pair $(\mathcal{M}_{t+1}, \mathcal{H}_{t+1})$.
\end{enumerate}

Every trace serves as both AEGIS diagnostic evidence and GRPO training signal. The harness evolution (step~4) and model update (steps~5--6) read the same buffer but neither conditions on the other's output within the same iteration; both must complete before the next rollout begins.

\subsection{Optimization Substrates}
\label{sec:coevolution-dual-grpo}

\paragraph{Harness side (non-parametric optimization).}
Harness evolution proceeds as in Section~\ref{sec:method-aegis}, drawing on the replay buffer $\mathcal{B}$ for trace evidence. The principal difference from standalone AEGIS is that $\mathcal{B}$ contains trajectories from multiple model checkpoints $\mathcal{M}_0, \mathcal{M}_1, \ldots, \mathcal{M}_t$, exposing the Digester to behavioral variation from both model updates and harness edits.

\paragraph{Model side (parametric optimization via GRPO).}
The key design choice is the \textit{cross-harness grouping criterion} (formalized in Section~\ref{sec:coevolution-grpo}): all trajectories sharing a task identifier form one GRPO group regardless of which harness or model checkpoint produced them, so that within-group variation reflects strategy differences rather than sampling noise alone.

\paragraph{Complementarity.}
The harness update makes discrete structural changes (adding a tool, replacing a control processor, restructuring the prompt) that cannot be expressed as parameter updates. The model update makes fine-grained behavioral adjustments (when to invoke which tool, how to phrase a query, when to terminate) that depend on high-dimensional in-context state and cannot be captured by symbolic specification. The harness defines coarse-grained strategy architecture; the model learns to exploit it.

\subsection{Model Training via Cross-Harness GRPO}
\label{sec:coevolution-grpo}

We adopt Group Relative Policy Optimization (GRPO)~\cite{shao2024deepseekmath}. Formally, each trajectory in the buffer is generated as:
\begin{equation}
    \tau_i \sim \text{Agent}(\mathcal{M}_k,\, \mathcal{H}_k,\, x_i), \quad k \in \{0, 1, \ldots, t\},
  \label{eq:trajectory-generation}
\end{equation}
where $i$ is the $(x, \tau)$ index in the buffer $\mathcal{B}$, $\mathcal{M}_k$ and $\mathcal{H}_k$ are the model checkpoint and harness used to roll out task $x_i$ into trajectory $\tau_i$. Because FIFO eviction bounds the buffer to recent iterations, buffered trajectories come from model versions close to the current policy. Yet they differ markedly in strategy (tool selection, prompt structure, control-flow logic), a diversity that stems from the successive harness versions $\mathcal{H}_0, \ldots, \mathcal{H}_t$. Unlike single-policy RL, where within-group variation reduces to stochastic sampling, here harness identity dominates that variation, which makes the cross-harness grouping criterion (Eq.~\ref{eq:cross-harness-group}) essential for meaningful advantage estimation.

Formally, for a task $x$, the trajectory group collects all traces of $x$ regardless of which $(\mathcal{M}_k, \mathcal{H}_k)$ pair produced them:
\begin{equation}
  \mathcal{G}_x = \{\tau_i \in \mathcal{B} \mid \text{task}(\tau_i) = x\} = \bigcup_{k} \{\tau \sim \text{Agent}(\mathcal{M}_k, \mathcal{H}_k, x)\}.
  \label{eq:cross-harness-group}
\end{equation}
The model therefore receives gradient signal from inter-strategy reward contrasts, rather than from stochastic variation within a fixed strategy alone, which enables it to internalize strategies that succeeded across harness versions.

\paragraph{Task-level alignment, not action-level.}
Cross-harness GRPO performs \textit{task-level} alignment: trajectories from different harness versions are grouped by task identity and compared by verifier reward alone. No action-level alignment is required, so harness versions with incompatible action spaces (different tool schemas, different prompt structures, different control-flow processors) coexist in the same group without conflict. When computing the policy gradient, each trajectory $\tau_i$ is replayed under the harness version $\mathcal{H}_k$ that produced it: the model's log-probabilities $\pi_\theta(\tau_i \mid x)$ are evaluated against the prompt, tool schema, and observation context that $\mathcal{H}_k$ would have constructed at each turn. The GRPO gradient thus operates entirely on model output tokens conditioned on harness-specific context, rather than on harness structural actions or environment transitions. This design decouples harness evolution (which may freely alter the action space across versions) from model training (which only requires token-level log-probabilities under each trajectory's own harness context).

The group-relative advantage is:
\begin{equation}
  \hat{A}(\tau_i) = \frac{r_i - \mu(\mathcal{G}_x)}{\sigma(\mathcal{G}_x) + \epsilon},
  \label{eq:grpo-advantage}
\end{equation}
where $r_i$ is the reward for trajectory $\tau_i$, and $\mu(\mathcal{G}_x)$, $\sigma(\mathcal{G}_x)$ are the within-group reward mean and standard deviation. The evolving harness acts as a structured exploration operator for the model's RL: each new version injects a distinct mode of behavior into the task's sampling distribution, and the advantage in Eq.~\ref{eq:grpo-advantage} commits the model toward whichever modes the verifier scores highest. The exploration breadth that single-policy sampling cannot provide is thus supplied by the evolving scaffold itself.

The policy objective to maximize is:
\begin{equation}
  \mathcal{J}_{\text{GRPO}}(\theta) = \mathbb{E}_{x,\, \tau_i \sim \mathcal{B}} \left[ \min\!\left( \rho_i(\theta)\,\hat{A}(\tau_i),\; \text{clip}\!\left(\rho_i(\theta),\, 1{-}\epsilon_c,\, 1{+}\epsilon_c\right)\hat{A}(\tau_i) \right) \right] - \beta\, D_{\text{KL}}\!\left(\pi_\theta \,\|\, \pi_{\text{ref}}\right),
  \label{eq:grpo-objective}
\end{equation}
where
\begin{equation}
  \rho_i(\theta) = \frac{\pi_\theta(\tau_i \mid x)}{\pi_{\theta_{\text{old}}}(\tau_i \mid x)}, \qquad \pi_{\theta_{\text{old}}} = \mathcal{M}_{d},
  \label{eq:grpo-ratio}
\end{equation}
is the importance-sampling ratio between the current policy $\mathcal{M}_{k}$ and the checkpoint $\mathcal{M}_{d}$ that generated $\tau_i$ (Eq.~\ref{eq:trajectory-generation}), $\epsilon_c$ is the clipping threshold, and $\beta\, D_{\text{KL}}(\pi_\theta \| \pi_{\text{ref}})$ penalizes divergence from the fixed reference model $\pi_{\text{ref}}$. The behavior policy $\pi_{\theta_{\text{old}}}$ in the ratio and the reference policy $\pi_{\text{ref}}$ in the KL term are distinct: $\pi_{\text{ref}} = \mathcal{M}_0$ is fixed throughout training, while $\pi_{\theta_{\text{old}}}$ varies per trajectory and must be recovered from the buffer (Section~\ref{sec:coevolution-offpolicy}).

\subsection{Off-Policy Training over a Mixed-Policy Buffer}
\label{sec:coevolution-offpolicy}

The replay buffer is intrinsically off-policy: at iteration $t$ it holds trajectories generated by checkpoints $\mathcal{M}_0, \mathcal{M}_1,\dots, \mathcal{M}_t$ under harnesses $\mathcal{H}_0, \mathcal{H}_1, \dots, \mathcal{H}_t$ (Eq.~\ref{eq:trajectory-generation}), so the buffer distribution does not match the policy $\pi_\theta$ under update. Recovering $\pi_{\theta_{\text{old}}}$ for each buffered trajectory is the central off-policy challenge.

\paragraph{Behavior policy $\pi_{\theta_{\text{old}}}$.}
The importance ratio (Eq.~\ref{eq:grpo-ratio}) corrects the gap between $\pi_\theta$ and the checkpoint $\mathcal{M}_{k}$ that produced $\tau_i$. Since $\mathcal{M}_{k}$ varies across the buffer, $\pi_{\theta_{\text{old}}}(\tau_i)$ cannot be recovered from any single model: we materialize it at buffer insertion via one forward pass under $\mathcal{M}_{k}$, cache the token-level log-probabilities on disk, and reuse them at every gradient step. This decouples the cached behavior log-probabilities from the current log-probabilities $\pi_\theta(\tau_i)$ recomputed each step.

\paragraph{Bounded off-policy bias.}
FIFO eviction caps the buffer at $C$ trajectories; with $s$ samples per round the maximum model-version lag is $\lfloor C/s \rfloor$ rounds, so every cached $\pi_{\theta_{\text{old}}}$ originates within a bounded window of $\pi_\theta$ and the policy that generated a trajectory never differs greatly from the one being updated. The same window bounds harness staleness, so the cross-harness groups (Eq.~\ref{eq:cross-harness-group}) mix only recent scaffold versions, and the model is never trained predominantly against an obsolete harness.

\paragraph{Replay reuse at no added rollout cost.}
The dominant cost of agentic RL is the rollout (executing the agent in the environment: model decoding, tool calls, and verification), not the gradient update. In co-evolution a single round of exploration produces one set of trajectories that serves both updates: the same traces drive the AEGIS harness update (Section~\ref{sec:method-aegis}) and, through the shared buffer (Section~\ref{sec:coevolution-loop}), the cross-harness GRPO model update. GRPO consumes these trajectories by replay and issues no rollouts of its own. The marginal cost of adding the model update is therefore confined to (i) one cached forward pass per trajectory to record $\pi_{\theta_{\text{old}}}$ and (ii) the gradient steps themselves, both of which are rollout-free. No trajectory is generated solely to train the model. Joint optimization is therefore economical: it buys model improvement for the price of offline training compute alone, without any rollouts beyond those harness evolution already performs.

\section{Experiments}
\label{sec:experiments}

We evaluate \ProjectName along five axes: overall effectiveness across benchmarks and model families (Section~\ref{sec:exp-main}), the impact of variant-management strategies on stability (Section~\ref{sec:exp-strategy}), the relative contribution of evolver architecture versus infrastructure (Section~\ref{sec:exp-meta-agent}), gains from joint model--harness co-evolution (Section~\ref{sec:exp-coevolution}), and empirical confirmation of the predicted failure modes (Section~\ref{sec:exp-failure}).

\subsection{Experimental Setup}
\label{sec:exp-setup}

\paragraph{Benchmarks.}
As summarized in Table~\ref{tab:benchmarks}, we evaluate on five benchmarks spanning multi-step retrieval, embodied planning, web interaction, multi-turn dialogue, and software engineering. Unless otherwise noted, each experiment runs for up to $T{=}15$ evolution rounds with early stopping after $P{=}3$ consecutive rounds without a shipped edit. The full task set is evaluated every round (no subsampling). The meta-agent token budget varies by benchmark (100M--175M total) but is held constant across task agents within a benchmark.

\begin{table}[!htbp]
\centering
\small
\caption{Benchmark characteristics.}
\label{tab:benchmarks}
\begin{tabular}{llcl}
\toprule
\textbf{Benchmark} & \textbf{Domain} & \textbf{Sampled Tasks} & \textbf{Verifier} \\
\midrule
GAIA (Level 1--3) & Multi-step retrieval & 103 & Exact match \\
ALFWorld & Embodied planning & 134 & Goal completion \\
WebShop & Web interaction & 100 & Attribute match \\
$\tau^3$-Bench & Multi-turn dialogue & 3 domains & Rule compliance \\
SWE-bench Verified & Software engineering & 55 & Patch resolution \\
\bottomrule
\end{tabular}
\end{table}

\paragraph{Models.}
We distinguish two roles: the \textit{meta-agent} (Claude Opus 4.6 unless otherwise noted) drives the AEGIS evolution loop; the \textit{task agent} runs under the evolved harness to solve benchmark tasks. Task agents span three families (Claude Sonnet 4.6, GPT-5.4, and Qwen3.5-9B) to test whether a single meta-agent can evolve effective harnesses across model families.

\paragraph{Baselines.}
\textbf{(1) Static Harness}: a \ProjectName configuration constructed from published benchmark-specific prompts and tool definitions, held fixed across all rounds.
\textbf{(2) Claude Code SDK (CC SDK)}\footnote{Claude Code SDK v0.0.25, \texttt{model="opus"} (Claude Opus 4.6), \texttt{max\_turns=200}. Experiments conducted in May 2026.}: a single-agent evolver (one LLM session per round) that replaces the four-stage pipeline while retaining the same infrastructure and round budget, isolating AEGIS's multi-stage architecture from the shared infrastructure (Section~\ref{sec:exp-meta-agent}). This baseline also serves as a proxy for monolithic evolvers such as SICA~\cite{robeyns2025self}.

\paragraph{Metrics.} Task success rate (\%) under the benchmark-specific verifier. Each task receives two independent attempts per round (pass@2: solved if either succeeds), reducing sampling noise while preserving a binary per-task signal for the seesaw constraint (at the cost of masking sub-threshold success-probability drift; Section~\ref{sec:exp-strategy}).

\paragraph{Scope.} All reported gains are measured on the same task set used for evolution; held-out generalization to unseen tasks is not evaluated in this work.

\begin{figure*}[!htbp]
  \centering
  \includegraphics[width=0.85\textwidth]{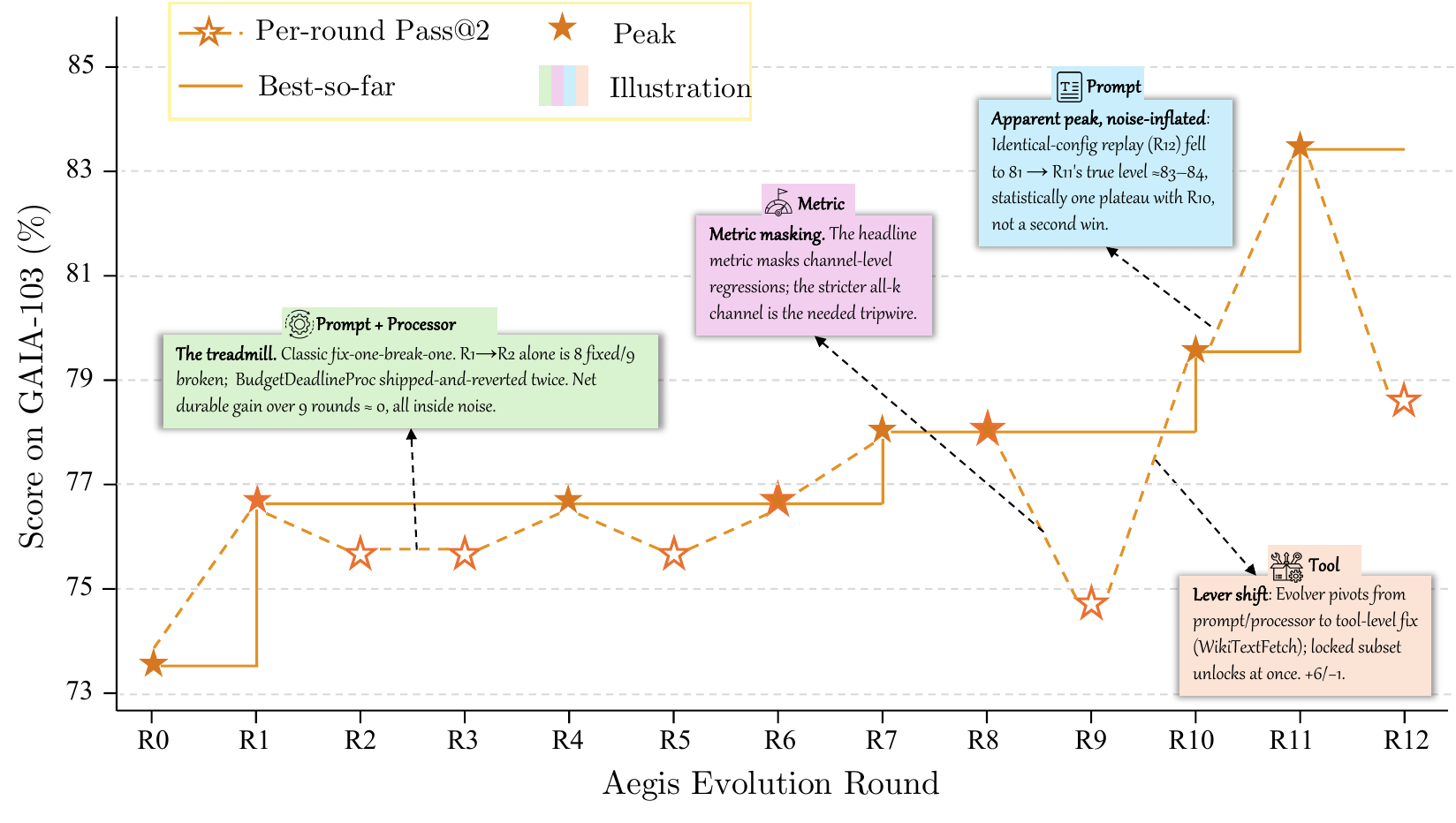}
  \caption{Evolution trajectories (pass@2 success rate vs.\ round). Dashed lines: static-harness baselines.}
  \label{fig:evolution-curves}
\end{figure*}

\subsection{Main Results}
\label{sec:exp-main}

Table~\ref{tab:main-results} and Figure~\ref{fig:evolution-curves} report pass@2 success rates before and after harness evolution. AEGIS improves 14 of 15 model--benchmark configurations, with an average gain of +14.5\% (up to +44.0\%). The single stagnating configuration (GAIA, GPT-5.4, $\Delta{=}0.0$) reflects a fundamental limitation of single-harness evolution on heterogeneous task sets; Section~\ref{sec:exp-strategy} shows that variant isolation resolves this. One configuration regressed mid-run ($\tau^3$-Bench Telecom, $-$14.0\% at R7) due to accumulated same-type edits, recovering by R9 (Section~\ref{sec:exp-failure}).

\begin{table}[!htbp]
\centering
\small
\caption{Main results (pass@2 success rate, \%). Evolved = peak accuracy achieved. ``--'' indicates domain-averaged results where no single peak round applies.}
\label{tab:main-results}
\begin{tabular}{llcccc}
\toprule
\textbf{Benchmark} & \textbf{Task agent} & \textbf{Initial} & \textbf{Evolved} & \textbf{$\Delta$} & \textbf{Best round} \\
\midrule
\multirow{3}{*}{ALFWorld}
  & Claude Sonnet 4.6 & 83.6 & 94.8 & +11.2 & 7 \\
  & GPT-5.4           & 76.9 & \textbf{97.8} & +20.9 & 4 \\
  & Qwen3.5-9B        & 53.0 & 97.0 & +44.0 & 9 \\
\midrule
\multirow{3}{*}{WebShop}
  & Claude Sonnet 4.6 & 60.0 & \textbf{76.0} & +16.0 & 7 \\
  & GPT-5.4           & 55.0 & 73.0 & +18.0 & 8 \\
  & Qwen3.5-9B        & 36.0 & 49.0 & +13.0 & 7 \\
\midrule
\multirow{3}{*}{GAIA}
  & Claude Sonnet 4.6 & 73.8 & \textbf{83.5} & +9.7 & 11 \\
  & GPT-5.4           & 73.8 & 73.8 & 0.0 & 4 \\
  & Qwen3.5-9B        & 20.3 & 37.4 & +17.1 & 4 \\
\midrule
\multirow{3}{*}{SWE-bench Verified}
  & Claude Sonnet 4.6 & 76.4 & \textbf{87.3} & +10.9 & 3 \\
  & GPT-5.4           & 45.5 & 63.6 & +18.2 & 3 \\
  & Qwen3.5-9B        & 23.6 & 41.8 & +18.2 & 2 \\
\midrule
\multirow{3}{*}{$\tau^3$-Bench (Avg.)}
  & Claude Sonnet 4.6 & 89.6 & \textbf{95.0} & +5.4 & -- \\
  & GPT-5.4           & 76.2 & 90.7 & +14.5 & -- \\
  & Qwen3.5-9B        & 93.5 & 94.6 & +1.1 & -- \\
\bottomrule
\end{tabular}
\end{table}

\textbf{Overall performance.} Evolution improves 14 of 15 configurations. Gains range from +11.2\% to +44.0\% on ALFWorld, +13.0\% to +18.0\% on WebShop, and +10.9\% to +18.2\% on SWE-bench Verified. On GAIA, Sonnet 4.6 (+9.7\%) and Qwen3.5-9B (+17.1\%) improve, while GPT-5.4 stagnates ($\Delta{=}0.0$; resolving its failures demands mutually conflicting edits that no single-harness strategy can accommodate). On $\tau^3$-Bench, GPT-5.4 gains most (+14.5\%) while Qwen3.5-9B gains only +1.1\% due to its near-ceiling 93.5\% baseline.

\textbf{Inverse scaling with baseline performance.} Across benchmarks, the weakest task agent (Qwen3.5-9B) consistently gains most: +44.0\% on ALFWorld (baseline 53.0\%), +17.1\% on GAIA (baseline 20.3\%), and +18.2\% on SWE-bench Verified (baseline 23.6\%). Stronger models (Sonnet 4.6, GPT-5.4) gain less on ALFWorld (+11.2\%, +20.9\%) and SWE-bench (+10.9\%, +18.2\%). The exception is GAIA GPT-5.4 ($\Delta{=}0.0$), where task heterogeneity prevents a single harness from improving aggregate accuracy---an observation that motivates the variant-isolation ablation in Section~\ref{sec:exp-strategy}. The overall pattern suggests that weaker models exhibit more behavioral gaps addressable by harness-level edits; once baseline performance is sufficiently high, remaining failures increasingly require task-specific adaptations rather than global improvements.

\textbf{Cross-model generalization.} The meta-agent (Opus 4.6) evolves harnesses for task agents across model families without family-specific adaptation. On ALFWorld, cross-family agents (GPT-5.4: +20.9\%, Qwen3.5-9B: +44.0\%) gain more than the same-family agent (Sonnet 4.6: +11.2\%), indicating that gain magnitude tracks baseline performance rather than proximity to the meta-agent's family.

\textbf{Convergence rate tracks failure-mode concentration.} ALFWorld (GPT-5.4) peaks at R4 and SWE-bench Verified (all agents) peaks at R2--R3; in both cases, failures concentrate in one or two component types, enabling rapid convergence. GAIA (Sonnet 4.6) requires 11 rounds because failures span four component types (prompt, tool, processor, configuration), forcing sequential exploration of multiple edit neighborhoods.

\paragraph{Domain-level variation within $\tau^3$-Bench.}
The averaged $\tau^3$-Bench gains mask substantial per-domain variation. GPT-5.4 gains +25.4\% on Telecom (67.5\%~$\to$~93.0\% at R2) and +9.7\% on Retail (84.2\%~$\to$~93.9\% at R6). However, Sonnet 4.6 on Telecom regresses $-$14.0\% in a single round (R7) due to accumulated same-type edits, recovering by R9 (Section~\ref{sec:exp-failure}). This illustrates a structural limitation of per-edit gating: sub-threshold coupling from consecutive same-type edits accumulates undetected until a tipping point triggers visible regression.

\paragraph{Post-peak degradation on SWE-bench.}
On SWE-bench Verified (GPT-5.4), evolution peaks at 63.6\% (R3, +18.2\%) but degrades to 50.9\% by R5 ($-$12.7\% from peak); final accuracy still exceeds the static baseline by +5.4\%. Two factors accelerate degradation on this benchmark: (1)~with only 55 tasks, each task flip shifts aggregate accuracy by ${\sim}$1.8\% (vs.\ ${\sim}$1.0\% at $n{=}103$), so fewer regressions suffice to produce visible decline; and (2)~structural code edits have a broader blast radius than prompt edits. This parallels the GAIA GPT-5.4 stagnation: both cases motivate the variant-isolation strategy evaluated in Section~\ref{sec:exp-strategy}.

\subsection{Evolution Strategy Comparison}
\label{sec:exp-strategy}

The main experiments (Table~\ref{tab:main-results}) use the Global strategy: a single harness evolved across all tasks. Table~\ref{tab:strategy} compares this default with a variant-isolation strategy on GAIA (103 tasks, GPT-5.4, 15 rounds, AEGIS evolver).

\begin{table}[!htbp]
\centering
\small
\caption{Evolution strategy comparison (GAIA, GPT-5.4, AEGIS evolver, 15 rounds). Final$-$Peak indicates stability; negative values signal catastrophic forgetting.}
\label{tab:strategy}
\begin{tabular}{lcccc}
\toprule
\textbf{Strategy} & \textbf{Final (\%)} & \textbf{Peak (\%)} & \textbf{Final$-$Peak} & \textbf{Tokens} \\
\midrule
Ensemble (up to $K$ variants) & \textbf{87.4} & 87.4 & 0.0  & 107.8M \\
Global (single harness)            & 49.5 & 73.8 & $-$24.3 & 143.7M \\
\bottomrule
\end{tabular}
\end{table}

\textbf{Failure mechanism of Global.} The Global strategy maintains a single harness for all 103 tasks. It peaks early at R4 (73.8\%) before degrading steadily: subsequent edits introduce sub-threshold regressions that are individually undetectable under pass@2's binary signal yet compound into aggregate decline. The peak--final gap ($-$24.3\%) far exceeds the per-round binomial 95\% confidence interval ($\pm$8.5\% at $n{=}103$, $p{\approx}0.74$), ruling out evaluation noise and confirming catastrophic forgetting (Section~\ref{sec:aegis-pathologies}). This explains the $\Delta{=}0.0$ stagnation for GAIA GPT-5.4 in Table~\ref{tab:main-results}: Global cannot sustain improvement on this heterogeneous task set.

\textbf{Why Ensemble prevents cross-variant forgetting.} Ensemble routing maintains up to $K$ harness variants and routes each task to the variant with the highest prior success rate. Edits are proposed and evaluated \textit{per-variant}, so an edit improving one cluster cannot regress another. The comparison confirms three predicted properties: (1)~non-degrading aggregate trajectory (peak = final), (2)~later peak (R14 vs.\ R4), indicating sustained productive exploration, and (3)~lower token consumption (107.8M vs.\ 143.7M), because each edit is evaluated only against its target cluster rather than the full task set, and edits target only their assigned cluster, avoiding the wasted proposals that accumulate when a degrading single harness is evaluated against all tasks.

\textbf{Summary.} Variant isolation resolves the stagnation observed under Global, lifting GAIA GPT-5.4 from $\Delta{=}0.0$ to +13.6\% (87.4\%, non-degrading). Finer-grained strategies (Domain-aware clustering, Task-level tournament) were explored at pilot scale (30--40 tasks, $\leq$8 rounds) but lack sufficient rounds and tasks for statistically meaningful comparison.

\subsection{Meta-Agent Effectiveness}
\label{sec:exp-meta-agent}

To disentangle evolver architecture from infrastructure, we replace the four-stage AEGIS pipeline with a single-agent CC SDK evolver that shares the same model (Opus 4.6), round budget, and infrastructure. Both evolvers run under variant isolation (introduced in Section~\ref{sec:exp-strategy}) to ensure non-degrading trajectories. Table~\ref{tab:meta-agent} reports the comparison on GAIA (103 tasks, GPT-5.4, 15 rounds).

\begin{table}[!htbp]
\centering
\small
\caption{Meta-agent architecture comparison (GAIA, GPT-5.4, variant isolation, 15 rounds). Both evolvers use Opus 4.6.}
\label{tab:meta-agent}
\begin{tabular}{lccc}
\toprule
\textbf{Evolver} & \textbf{Accuracy (\%)} & \textbf{Best round} & \textbf{Tokens} \\
\midrule
AEGIS   & \textbf{87.4} & R14 & 107.8M \\
CC SDK  & 86.4 & R12 & 123.1M \\
\bottomrule
\end{tabular}
\end{table}

\textbf{Accuracy is comparable; efficiency differs.} The 1.0\% accuracy gap falls within one standard error (${\sim}$3.3\% at $n{=}103$), indicating that the four-stage decomposition does not improve final accuracy at this meta-agent capability level. However, the single-agent variant consumes ${\sim}$14\% more tokens (123.1M vs.\ 107.8M). We attribute this to the Digester's compression: it reduces ${\sim}$10M raw trace tokens to ${\sim}$10K structured summaries before downstream stages consume them. Without this stage, the single-agent evolver must truncate traces to fit its context window, yielding less-informed edits that are rejected by the gate more frequently, wasting tokens on failed proposals.

\textbf{Implication.} With a capable meta-agent under variant isolation, accuracy gains derive primarily from \ProjectName's infrastructure (typed components enabling isolation, structured traces enabling diagnosis) rather than the evolver's internal architecture. The four-stage decomposition contributes efficiency (${\sim}$12\% fewer tokens) and interpretability (auditable intermediate artifacts) but not measurable accuracy at this scale.

\subsection{Co-Evolution}
\label{sec:exp-coevolution}

\begin{figure*}[!htbp]
  \centering
  \includegraphics[width=0.75\textwidth]{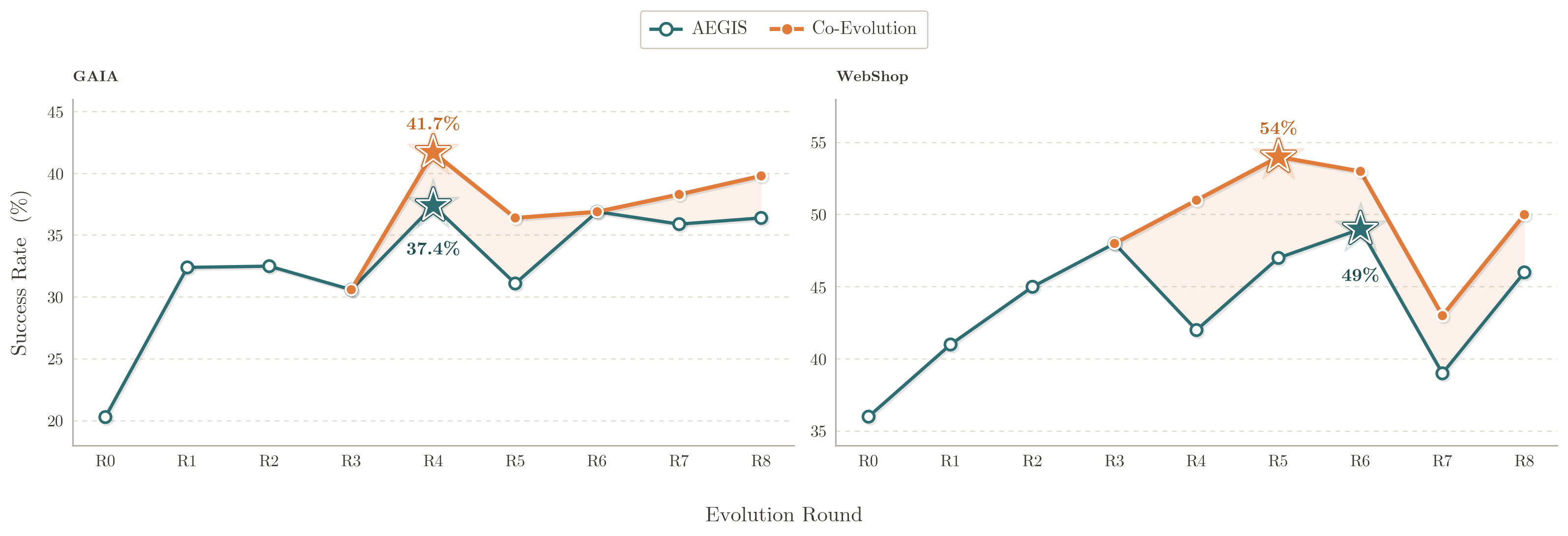}
  \caption{Co-evolution vs.\ harness-only evolution (AEGIS, model frozen) on GAIA and WebShop. Stars mark each method's peak; the shaded band is the co-evolution gain.}
  \label{fig:coevolution-curves}
\end{figure*}

This experiment tests whether interleaving harness evolution with model RL (Section~\ref{sec:method-coevolution}) yields gains beyond harness-only evolution. As shown in Figure~\ref{fig:coevolution-curves}, we compare the two regimes on GAIA and WebShop using a Qwen3.5-9B task agent. Both conditions share a fixed-capacity FIFO replay buffer: each round runs the current agent on the adaptation batch, a fixed verifier scores the resulting traces, and both harness evolution (AEGIS) and model training (cross-harness GRPO) update over the same buffer (Section~\ref{sec:coevolution-loop}). Section~\ref{sec:method-coevolution} predicts that each single-optimization route stalls at its own ceiling: harness-only at the \textit{scaffolding ceiling}, model-RL-only at the \textit{training-signal ceiling}. Co-evolution addresses both ceilings by enabling the model to internalize strategies that successive harness versions introduce.

\textbf{Experimental setup.} We run both regimes on the GAIA text-only subset (103 tasks) and a WebShop subset (100 tasks) with a Qwen3.5-9B task agent. GAIA exercises live web tools whose latency and availability fluctuate, so each round is evaluated twice and averaged. Both subsets are small, so we set the optimizer batch to the entire replay buffer and size the buffer as a four-round sliding window: at two rollouts per task this is 824 traces on GAIA ($103\times2\times4$) and 400 on WebShop ($100\times1\times4$), which supplies enough within-group samples for GRPO to estimate advantages stably. Training uses learning rate $1\times10^{-6}$, GRPO clip $\epsilon=0.2$, no KL penalty (coefficient $0$), and 5 training steps per round. The GAIA agent is equipped with web search (Baidu API), web fetch, bash, and file read; WebShop uses its environment's built-in action tools. Rewards are $0.9\times$correctness plus $0.1\times$format on GAIA, and WebShop's native attribute-match reward (a task passes only at reward $=1.0$).

\textbf{Co-evolution exceeds harness-only evolution.} As Figure~\ref{fig:coevolution-curves} shows, interleaving cross-harness GRPO with harness evolution over a shared replay buffer raises peak success on both benchmarks: GAIA 37.4\%~$\to$~41.7\% (+4.3\%) and WebShop 49.0\%~$\to$~54.0\% (+5.0\%), averaging +4.7\% over the model-frozen baseline. The two curves coincide until joint training takes effect (R4), then diverge, with co-evolution at or above harness-only for the remainder of the run. The gap persists to the final round (GAIA 36.4\%~$\to$~39.8\%, WebShop 46.0\%~$\to$~50.0\%) and is wider on WebShop, where more room remains for model-level improvement beyond the harness-only plateau. Co-evolution thus lifts end-of-run accuracy, not merely the peak.

\textbf{Co-evolution breaks the scaffolding ceiling.} Harness-only evolution plateaus at ${\sim}$37\% on GAIA and ${\sim}$49\% on WebShop. Co-evolution clears these plateaus: cross-harness GRPO enables the model to internalize strategies from successive harness versions, so later edits build on learned behavior rather than compensating for a fixed model's intrinsic limitations.

\subsection{Failure Analysis}
\label{sec:exp-failure}

\begin{figure}[!t]
  \centering
  \includegraphics[width=0.7\textwidth]{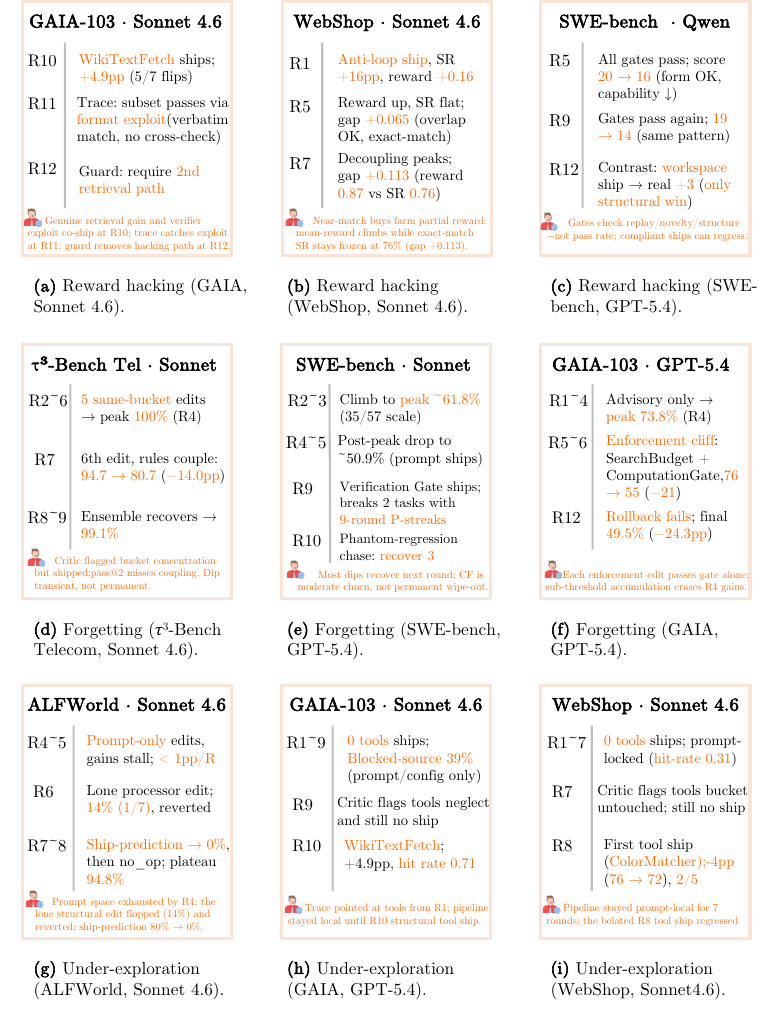}
  \caption{Failure cases organized by pathology (rows: reward hacking, catastrophic forgetting, under-exploration).}
  \label{fig:failure-cases}
\end{figure}

We present three case studies, one per pathology predicted by the operational mirror (Section~\ref{sec:aegis-pathologies}): reward hacking, catastrophic forgetting, and under-exploration. For each case we document the detection signal that first surfaced the issue, the root cause identified through trace analysis, and the outcome---whether the pipeline self-corrected or required manual intervention. Figure~\ref{fig:failure-cases} provides the full set of confirmed and pending cases organized by pathology type.

\paragraph{Reward hacking (GAIA, Sonnet 4.6, R10).}
At R10, the pipeline shipped a composite edit (tool + prompt + configuration) whose manifest predicted improved retrieval. The edit passed the seesaw constraint and raised accuracy from 74.8\% to 79.6\%. Trace analysis at R11 revealed that the tool genuinely fixed retrieval for most newly passing tasks, but a subset passed by exploiting format regularities in the verifier rather than performing actual retrieval. The Planner flagged this pathway at R12, and the resulting edit introduced a guard restricting the tool to tasks whose output could be cross-checked against a second retrieval path.

\paragraph{Catastrophic forgetting ($\tau^3$-Bench, Sonnet 4.6, Telecom, R7).}
Evolution on Telecom shipped same-type prompt/processor edits across five consecutive rounds (R2--R6), each appending a ``reminder'' rule. Compliance rose from 89.5\% to 100\% at R4, then regressed to 94.7\% by R6 as later rules conflicted with earlier ones. The R7 Critic flagged the concentration risk (``All 5 prior ships occupy the same bucket: [prompt, processor]'') but still approved the edit for shipping because ship-prediction accuracy remained high (R2--R6: 23/24, 5/6, 4/5, 7/7, 2/3) and no regressions were recorded. The sixth reminder degraded compliance from 94.7\% to 80.7\% ($-$14.0\%) via cross-rule conflicts that destabilized previously passing tasks. This regression evaded the seesaw constraint because pass@2 registers only per-task binary flips, not sub-threshold coupling. The pipeline self-corrected by R9 once the Planner diagnosed the concentration pattern and proposed a structural edit that replaced the conflicting reminder stack.

\paragraph{Under-exploration (ALFWorld, Sonnet 4.6, R4--R7).}
Between R4 and R7, the pipeline shipped predominantly prompt-level edits, yielding $<$1\% gain per round. Ship-prediction accuracy (the fraction of manifest-predicted task flips that materialize) dropped from 80\% (R3) to 0\% (R7), signaling prompt-space exhaustion. The sole structural edit in this window (a processor-level change at R6) achieved only 14\% ship-prediction accuracy (1/7 predicted flips materialized), suggesting that the Planner lacked sufficient structural-edit history to calibrate hypotheses beyond the prompt neighborhood.

\paragraph{Summary.}
All three pathologies predicted by the operational mirror appear in practice. The pipeline detected and mitigated reward hacking within two rounds (R10--R12). Decaying ship-prediction accuracy diagnosed under-exploration (R4--R7). The catastrophic-forgetting case exposes a structural limitation of per-edit gating: sub-threshold coupling accumulates undetected until it exceeds the per-task detection threshold (Telecom R7). On $\tau^3$-Bench Telecom, the pipeline self-corrected (R8--R9) because the failure was localized to one domain; on GAIA (GPT-5.4), the same mechanism produces sustained stagnation ($\Delta{=}0.0$) because conflicting edits prevent any net gain. Section~\ref{sec:exp-strategy} shows that variant isolation resolves this by confining edits to task-specific clusters.

\section{Discussion}
\label{sec:discussion}

\subsection{Why Compositional Structure Matters for Evolution}

As Table~\ref{tab:strategy} shows, the Global strategy (used in all main experiments) peaks early at 73.8\% (R4) on GAIA before collapsing to 49.5\% (peak--final gap: $-$24.3\%). Global uses \ProjectName's typed components but does not leverage them for isolation; every edit is evaluated against all tasks jointly. Under pass@2, a task whose success probability has degraded can still register as ``solved,'' so sub-threshold regressions evade the seesaw constraint. Preventing this collapse requires variant isolation, which composability enables: \ProjectName's compositional structure makes the \emph{intended scope} of each edit explicit, a precondition for variant isolation to confine each edit's evaluation to its target cluster rather than evaluating against the full task set indiscriminately (Section~\ref{sec:exp-strategy}).

The relationship parallels type systems: types do not generate correct programs, but they make incorrect programs \emph{detectable}. Analogously, typed components do not prevent bad edits, but make their scope \emph{explicit}, enabling independent variation. The strategy comparison suggests that variant isolation is necessary for stable evolution (Global, which lacks it, degrades after peaking); without compositional structure, the intended scope of an edit is undefined, making variant isolation ill-posed. Compositional structure does not, however, guarantee bounded behavioral effects: the $\tau^3$-Bench Telecom failure demonstrates that accumulated same-type edits can induce sub-threshold coupling that degrades multiple dialogue patterns simultaneously.

\subsection{The Role of Trace Richness}

\ProjectName's full execution trace $\tau$ provides diagnostic information beyond a scalar reward. The case studies (Section~\ref{sec:exp-failure}) confirm this: detecting reward hacking on GAIA (shipped at R10, detected at R11) required inspecting \emph{how} improvement occurred (format exploitation vs.\ genuine retrieval), and detecting under-exploration on ALFWorld (R4--R7) required tracking edit-type distribution and ship-prediction accuracy. Neither signal is recoverable from per-task binary outcomes alone.

These observations motivate a design principle: \textit{the richness of the feedback signal bounds the sophistication of evolution that can be safely performed}. From scalar reward alone, none of the three pathologies is detectable: a score change cannot distinguish reward hacking from genuine improvement, under-exploration from convergence, or catastrophic forgetting from evaluation noise. Trace structure makes each pathology diagnosable, provided prior-round traces exist for comparison. The $\tau^3$-Bench Telecom failure illustrates the boundary: despite five rounds of prior traces (R2--R6), accumulated regressions evaded the seesaw constraint because no individual edit crossed the detection threshold. Structured trace recording is therefore necessary for detecting pathologies, but not sufficient for preventing them: when coupling accumulates below the per-task detection threshold, traces record the symptoms only after damage has occurred.

\subsection{Scope and Limits of the Operational Mirror}

The RL--symbolic-space mirror is a design heuristic, not a formal framework. Classical RL convergence guarantees require sufficient exploration of the state--action space, a condition unattainable when states are symbolic harness configurations and actions are open-ended code edits. Under the Global strategy, GAIA (GPT-5.4) stagnates entirely ($\Delta{=}0.0$ over 15 rounds); the variant-isolation ablation (Section~\ref{sec:exp-strategy}) recovers stable improvement (87.4\% final = peak), but nothing guarantees this extends to longer horizons (where variants may over-specialize) or to task distributions whose inter-task dependencies prevent clean variant separation. The mirror also does not predict \textit{which} pathology will dominate: on $\tau^3$-Bench Telecom, catastrophic forgetting surfaced at R7; on ALFWorld, under-exploration dominated R4--R7; on GAIA, reward hacking surfaced only at R10.

We therefore treat the mirror as a design checklist rather than a predictive theory: it identifies failure modes to defend against but does not predict their ordering, timing, or relative severity. The three pathologies are representative, not exhaustive; additional RL phenomena (e.g., distribution shift when the adaptation batch diverges from deployment tasks, reward sparsity on hard benchmarks) may manifest as analogous failure modes in symbolic space.

\subsection{Generalization Across Model Families}

On ALFWorld, the Opus 4.6 meta-agent evolves harnesses for task agents from three model families:
\begin{itemize}
  \item Sonnet 4.6 (same family): 83.6\% $\to$ 94.8\% (+11.2\%)
  \item GPT-5.4 (different family): 76.9\% $\to$ 97.8\% (+20.9\%)
  \item Qwen3.5-9B (different family, weaker): 53.0\% $\to$ 97.0\% (+44.0\%)
\end{itemize}
The inverse-scaling effect (Section~\ref{sec:exp-main}) explains the magnitude ordering: gains track inverse baseline performance (Qwen $>$ GPT $>$ Sonnet) rather than proximity to the meta-agent's model family. All three configurations hold the meta-agent fixed (Opus 4.6) while varying the task agent; we do not evaluate whether a weaker meta-agent can achieve comparable gains.

A complementary ablation (Section~\ref{sec:exp-meta-agent}) finds that a single-agent evolver achieves comparable accuracy to the four-stage AEGIS pipeline (86.4\% vs.\ 87.4\%, within sampling noise at $n{=}103$) when both share the same meta-agent model and infrastructure. This suggests that at this meta-agent capability level, the four-stage decomposition primarily provides efficiency gains (${\sim}$12\% fewer tokens) and auditability rather than measurable accuracy improvement.

\subsection{Cost-Performance Tradeoffs}

As Table~\ref{tab:cost-summary} details, evolution incurs upfront compute that amortizes over subsequent task invocations.

\begin{table}[!htbp]
\centering
\small
\caption{Evolution cost summary. All main experiments use the Global (single-harness) strategy; the variant-isolation row is from the strategy ablation (Section~\ref{sec:exp-strategy}).}
\label{tab:cost-summary}
\begin{tabular}{lccc}
\toprule
\textbf{Experiment} & \textbf{Rounds} & \textbf{Total Tokens} & \textbf{Gain} \\
\midrule
GAIA, GPT-5.4 (Global)             & 15 & 143.7M & 0.0\% (peak $=$ initial) \\
GAIA, GPT-5.4 (Variant isolation, ablation) & 15 & 107.8M & +13.6\% \\
ALFWorld, Sonnet 4.6 (Global)       & 7  & 43.4M  & +11.2\% \\
\bottomrule
\end{tabular}
\end{table}

The strategy ablation (Section~\ref{sec:exp-strategy}) shows that variant isolation is both more effective (87.4\% vs.\ 49.5\% final) and more efficient (107.8M vs.\ 143.7M tokens) than Global on GAIA. The token reduction has two sources: (1)~structurally, each edit under variant isolation is evaluated only against its target cluster rather than the full task set, reducing per-round evaluation cost; (2)~under Global, the steadily degrading baseline causes more candidates to fail gating, wasting tokens on candidates that never ship. On benchmarks where evolution converges quickly (ALFWorld R4--R7, SWE-bench R2--R3), Global suffices and degradation does not materialize within the run horizon.

The evolved harness also affects \textit{per-task inference cost}. On GAIA, per-task token consumption drops by ${\sim}$25\% (targeted tool selection shortens trajectories); on ALFWorld, it rises by ${\sim}$60\% (task-decomposition prompts lengthen execution).

At deployment, the evolved harness is a static artifact requiring no meta-agent inference; tasks outside the evolution set are routed to the variant with the highest overall success rate on the evolution set. On GAIA, the upfront 107.8M tokens amortize within ${\sim}$1{,}300 invocations (${\sim}$83K tokens saved per invocation). On ALFWorld, per-task cost increases; the return is accuracy (+11.2\%), not cost reduction.

\subsection{Ethical Considerations}

Self-evolving agent systems require explicit oversight. \ProjectName provides three mechanisms:
\begin{enumerate}
  \item \textbf{Auditability}: every shipped edit carries a manifest and a rollback target; rejected candidates are archived with rejection reasons.
  \item \textbf{Deterministic gating}: the seesaw constraint rejects any edit that regresses even a single previously solved task under pass@2.
  \item \textbf{Human-in-the-loop}: the gating layer supports human approval for edits exceeding a configurable risk threshold (not exercised in our automated experiments).
\end{enumerate}
The $\tau^3$-Bench failure (Section~\ref{sec:exp-failure}) illustrates their limits: five consecutive same-type edits (R2--R6) accumulated sub-threshold coupling undetected by the seesaw constraint; the sixth edit (R7) triggered a visible $-$14.0\% regression, yet no individual edit violated the constraint. This is a structural limitation of per-edit gating: sub-threshold regressions accumulate undetected regardless of how many prior rounds have demonstrated apparent stability under the same constraint.

\subsection{Limitations}

Beyond the limitations noted above, five additional constraints bound the generality of our results:
\begin{itemize}
  \item \textbf{No held-out evaluation.} All reported gains are measured on the same task set used for evolution. Since we report peak accuracy and evaluate on the adaptation set itself, the numbers carry both selection bias and potential overfitting. Generalization to unseen tasks within the same distribution is plausible but untested.
  \item \textbf{Discrete action spaces only.} All experiments use agents with discrete, text-based action spaces. We have not tested whether the framework extends to continuous action spaces (e.g., robotic control).
  \item \textbf{Closed-source meta-agent.} AEGIS requires a meta-agent capable of multi-file code generation, structured trace analysis, and multi-step planning. Open-weight models approaching this capability level (e.g., Qwen3.5-72B, Llama-4-Maverick) remain untested as meta-agents.
  \item \textbf{Joint control assumption.} Co-evolution requires joint control over both harness evolution and model training. In practice, these concerns are often separated across teams or organizations, making a shared replay buffer (Section~\ref{sec:coevolution-loop}) impractical without cross-team coordination.
  \item \textbf{Benchmark coverage.} All SWE-bench Verified runs use a 55-task subsample, and $\tau^3$-Bench evaluates only three domains (Retail, Airline, Telecom). Conclusions, particularly the inverse-scaling effect, may not generalize to domains with different task heterogeneity or to larger evaluation sets.
\end{itemize}

\section{Conclusion}
\label{sec:conclusion}

We present \ProjectName, a composable runtime foundry that treats the harness as a first-class interface between model and environment. This interface can be composed from typed primitives, evolved from execution traces, and coupled with model training in a unified improvement loop. Across five benchmarks and three model families, \ProjectName achieves gains up to $+44.0$\% (average $+14.5$\% across 15 configurations) through trace-driven evolution over a compositional substrate, with co-evolution adding +4.7\% beyond harness-only evolution on two benchmarks. These results suggest that agent progress need not rely on model scaling alone: composing and evolving the runtime interface from execution feedback is a complementary and actionable lever, particularly for capability-limited agents where harness-level gains are largest.

\clearpage
\bibliographystyle{plainnat}
\bibliography{references}

\clearpage
\section*{Contributions and Acknowledgments}
\label{sec:contributions}

\vspace{4pt}
\noindent
\begin{minipage}[t]{0.45\textwidth}
\textbf{\textsf{\xiaomievblue{Core Contributors}}}
\begin{itemize}[leftmargin=12pt, itemsep=1pt, parsep=0pt, topsep=4pt]
  \item Tingyang Chen*
  \item Shuo Lu*
  \item Kang Zhao*
  \item Weicheng Meng
  \item Kun Shao\textsuperscript{\dag}
  \item Jian Luan\textsuperscript{\dag}
\end{itemize}
\end{minipage}%
\hfill
\begin{minipage}[t]{0.45\textwidth}
\textbf{\textsf{\xiaomievblue{Contributors}}}
\begin{itemize}[leftmargin=12pt, itemsep=1pt, parsep=0pt, topsep=4pt]
  \item Hanlin Teng
  \item Tianhao Li
  \item Chao Li
  \item Xule Liu
  \item Jian Liang
  \item Zhizhong Zhang
  \item Yuan Xie
  \item Heng Qu
\end{itemize}
\end{minipage}

\let\thefootnote\relax\footnotetext{* Equal Contribution \quad \textsuperscript{\dag} Corresponding Author}

\clearpage
\phantomsection
\addcontentsline{toc}{section}{Appendix}
\addtocontents{toc}{\protect\setcounter{tocdepth}{-1}}
\beginappendix
\newcommand{\tbd}[1]{\textcolor{red}{#1}}

\section{Experimental Setup: Full Details}
\label{app:setup}

This appendix expands the condensed setup of Section~\ref{sec:exp-setup} with
the full benchmark descriptions, the formal metric definitions, the evolution
protocol hyperparameters, and the runtime infrastructure.

\subsection{Benchmarks}
\label{app:benchmarks}

We evaluate on five benchmarks chosen to span the failure modes that harness
design most affects, from short-horizon embodied planning to long-horizon
software engineering.

\paragraph{GAIA.} The GAIA benchmark~\cite{mialon2024gaia} poses real-world questions that are conceptually simple for humans but require an agent to compose multiple actions (web search, file extraction, multimodal interpretation, arithmetic) and evaluates via exact match against a reference answer. This benchmark stresses open-ended tool-based reasoning, where the harness dictates how evidence is collected and synthesized.

\paragraph{ALFWorld.} The ALFWorld benchmark~\cite{shridhar2020alfworld} involves embodied instruction following where a text-based agent commands a simulated robotic agent in household settings. Given a natural-language goal (e.g., ``Put a cooled apple in the microwave''), the agent navigates rooms, identifies objects, and manipulates them via textual actions; performance is measured by goal-completion rate. This benchmark stresses multi-step planning and grounded search under a tight step budget. We use the 134 tasks from the valid-unseen set, spanning six task types: pick-and-place, pick-two-and-place, look-at-in-light, and three transform-then-place variants (heat, cool, clean).

\paragraph{WebShop.} WebShop~\cite{yao2022webshop} is a web-interaction benchmark in which an agent acts as a customer in a simulated online store. Given a textual product description, the agent must search, browse product pages, select the best-matching item, and purchase it; scoring reflects how well the chosen product satisfies the request. We evaluate on 100 instances sampled with a fixed seed, each run as an independent shopping session.

\paragraph{$\tau^3$-Bench.} $\tau^3$-Bench~\cite{yao2024tau} is a multi-turn dialogue benchmark in which the agent plays a customer-service assistant that must satisfy a user request while obeying an explicit domain policy. Performance is measured by rule compliance across the full conversation. The benchmark stresses dialogue-policy adherence: the harness must prevent the agent from agreeing to disallowed actions across many turns. For evaluation, we select three domains from the benchmark: Retail, Airline, and Telecom.

\paragraph{SWE-bench Verified.} SWE-bench Verified~\cite{jimenez2024swe} is a human-validated subset of SWE-bench in which each task requires an agent to resolve a real GitHub issue by editing the corresponding repository so that the project's hidden test suite passes. This benchmark stresses repository-level code editing: navigating a large codebase, localizing the relevant fault, implementing a patch, and avoiding regressions in existing tests. For evaluation, we sample a 55-task subset from SWE-bench Verified and measure performance by patch resolution.

\subsection{Evaluation-Set Design}
\label{app:evalset}

The sampled-task counts in Table~\ref{tab:benchmarks} denote the fixed evaluation sets scored at each evolution round. GAIA uses a fixed 103-task set drawn across the three difficulty levels (39/52/12). ALFWorld uses all 134 tasks from the valid-unseen split. WebShop uses 100 tasks randomly sampled from the dataset with a fixed seed, with each task run as an independent shopping session. For $\tau^3$-Bench, we select three domains (Retail, Airline, and Telecom) and score the full task list within each selected domain. For software engineering, we use a 55-task subset sampled from SWE-bench Verified. The same evaluation set for each benchmark is re-scored at every round, so the curves in Appendix~\ref{app:additional} measure round-over-round changes on fixed task sets rather than on moving samples.

\subsection{Metric Definitions}
\label{app:metrics}

\paragraph{Pass@$k$.}
For a configuration evaluated on a task set $D$ with $n$ rollouts per task, let
$r_{i,j}\in\{0,1\}$ denote the binary outcome of rollout $j$ on task $i$. Let
$c_i=\sum_{j=1}^{n} r_{i,j}$ be the number of successful rollouts for task $i$.
We report pass@$k$ using the standard unbiased estimator, i.e., the probability
that at least one of $k$ sampled rollouts solves the task:
\begin{equation}
\mathrm{Pass@}k
= \frac{1}{|D|}\sum_{i=1}^{|D|}
\left(
1 - \frac{\binom{n-c_i}{k}}{\binom{n}{k}}
\right).
\label{eq:passk}
\end{equation}
All evolution curves use pass@2 as the primary metric: each task receives two independent rollouts and is solved if either succeeds. This reduces sensitivity to single-rollout stochasticity while preserving a strict task-level criterion. Rollouts terminated by infrastructure failures (sandbox crashes, API timeouts) count as failures rather than being excluded, keeping results comparable to official leaderboard protocols.

\subsection{Evolution Protocol and Hyperparameters}
\label{app:protocol}

The hyperparameters used in our evolutionary algorithm are detailed in Table~\ref{tab:hyperparams}. In round 0, the baseline is a competent composed harness augmented with the benchmark-specific tool registry, rather than a minimal default. The plots therefore show gains relative to a competent initial harness. The meta-agent is Opus 4.6 across all experiments; task agents vary: Sonnet 4.6, GPT-5.4, and Qwen3.5-9B. Per-task step limits are determined by the benchmark in question since the interaction length varies greatly among tasks.

\begin{table}[!htbp]
\centering
\small
\caption{Evolution-protocol hyperparameters.}
\label{tab:hyperparams}
\begin{tabular}{lll}
\toprule
\textbf{Symbol} & \textbf{Meaning} & \textbf{Value} \\
\midrule
$K_t$ & candidates proposed per round & 4 \\
seeds & random seeds per cell & 3 \\
noise threshold & ignored single-round pass-count delta & $\pm$5\% \\
$\mathcal{H}_0$ & round-0 harness & Handcrafted base harness \\
meta-agent & drives Digester / Planner / Evolver / Critic & Opus 4.6 \\
task agent & model executing benchmark tasks & Sonnet 4.6; GPT-5.4; Qwen3.5-9B \\
concurrency & parallel task rollouts & 10 \\
\midrule
\multirow{5}{*}{max-steps}
  & GAIA & 20 \\
  & WebShop & 20 \\
  & ALFWorld & 15 \\
  & $\tau^3$-Bench & 200 \\
  & SWE-bench Verified & 200 \\
\bottomrule
\end{tabular}
\end{table}

\subsection{Runtime Infrastructure}
\label{app:runtime}

Every rollout runs inside a fresh environment instance re-attached per task, so
that side-effects (a WebShop cart, an ALFWorld world state, a shell working
directory) cannot leak between tasks. The runtime records each rollout's full
trajectory (every model call, tool call, and environment observation) to the
observability layer that the Digester subsequently compresses; the
cross-round ledgers are aggregated from this log. Task
rollouts execute at concurrency~10. The meta-agent runs at concurrency~4 with a 200-step limit per role. Co-evolution model training uses $8\times$\,H100 GPUs with batch size~256 and learning rate $1\times10^{-6}$.

\section{Prompts and Harness Defaults}
\label{app:prompts}

This appendix reproduces the prompts that drive the AEGIS outer loop and the
Round-0 task-agent defaults. The blocks below are the literal contents of the
corresponding files in the repository as of the commit that produced the
experiments in Section~\ref{sec:experiments}.

\subsection{Meta-Agent Prompts}
\label{app:meta-prompts}

\begin{promptfile}{planner/system\_prompt.md}
# Planner -- Round {{ round }}

Your goal: write a single `landscape.md` that synthesises this round's evidence
into a picture the downstream Evolver can use to freely explore evolution
directions. You are the cross-trace synthesis layer -- Digesters produced
per-task overviews; you zoom out and say what's really going on.

## What the landscape should convey

- Recurring failure modes across this round's digests -- your own grouping, not
  forced by exact-string matching.
  
- What was tried in previous rounds (journal.md, data/ship_outcomes.json,
  data/rejected_candidates.jsonl, archive/) and which outcomes held up.
  
- Tasks that persistently failed across rounds (data/task_history.jsonl) and
  theories about them that have NOT been tried yet.
  
- Whether last round's ship caused regressions. Read R{{ round }}/regressions.md
  first: it is the deterministic, k-aware list of tasks whose pass-state
  worsened versus the previous round, with the joint-suspect ships from
  R{{ round_minus_1 }} attached. The hit-rate in ship_outcomes.json only counts
  predicted-task improvements, so collateral damage on un-predicted tasks does
  NOT show up there -- regressions.md is the only place it surfaces. If the file
  lists regressions, put them at the top of the landscape (a dedicated
  "## Regressions to address" section), name the responsible ship's bucket(s),
  and flag each in unattempted_directions so the Evolver treats them as
  first-class targets.
  
- What the reputation signal says about which mutation layers have historically
  yielded (proposed -> shipped, window): {{ reputation_summary }}
  
- What scoreboard.json and ship_outcomes.json say about per-bucket hit rates. If
  one bucket has shipped 3+ rounds running with flat or declining hit rate while
  another has never been tried AND the digests point at failures it could
  address, say so: name the neglected bucket and the cluster it would target.
  Do not prescribe WHICH mutation to pick -- point at the evidence and let the
  Evolver decide the shape.
  
{% if round >= 2 %}- Prior Critic's strategy_concern, if any. Read
  R{{ round_minus_1 }}/decision.md; if its frontmatter has a non-empty
  strategy_concern, surface it at the TOP of the landscape, quoted verbatim,
  then note whether this round's evidence still supports it. The Critic writes
  strategy_concern to reach next round's Evolver, but the Evolver only reads
  landscape.md -- you are the relay.
{% endif %}

Be evidence-anchored. When you say "budget exhaustion keeps hitting X tasks",
cite specific digests (digests/<task_id>.md) or trajectory anchors
(trajectories/<task>_r0.jsonl#step_N). The Evolver will read what you cite.

Be selective, not exhaustive. Three coherent directions -> list three. One
overwhelming signal -> say so. Your reader decides how many candidates to build;
it benefits from clarity, not volume.

## Where evidence lives

Run root has INDEX.md, a catalog. Typical sources: overview.md (this round's
digests + patterns); digests/<task_id>.md (per-task analysis with anchors);
journal.md (prior memos); data/*.jsonl and data/ship_outcomes.json (cross-round
ledgers); archive/ (non-shipped manifests). No required reading list -- pull
what supports the synthesis.

## Output

One file via `write_tool`. The body is open-ended markdown; the only structural
expectation is a short YAML frontmatter so the Evolver can find your key
conclusions:

---
round: {{ round }}
top_themes:              # your synthesis, free-text tags
  - <theme-1>
persistent_failures:     # task_ids failed across >=2 rounds
  - <task_id>
unattempted_directions:  # approaches not tried yet per ship_outcomes
  - <short description>
---

## Landscape
<Open narrative. Evidence citations throughout.>
\end{promptfile}

\begin{promptfile}{evolver/system\_prompt.md}
# Evolver -- Round {{ round }}

Your goal: produce concrete evolution candidates whose shipping will raise next
round's benchmark pass rate. You decide how many candidates (K >= 1) -- one
high-value candidate beats three speculative ones, but if two genuinely
different directions both have strong evidence, produce both. Every candidate
must be evidence-driven with citations to raw traces or digests.

## Your stance

This role is research, not maintenance. Your value is in creative, rigorous,
breakthrough-level thinking -- not in iterating on the bucket the pipeline has
shipped most recently. When evidence points to a structural lever the harness
has never touched -- a new tool, a runloop parameter, a different processor-hook
time point -- propose it, even when the bucket has an empty reputation. Do not
let bucket history, gate-rejection fear, or implementation discomfort narrow
your search. Follow the evidence.

[... strategy-concern relay and revert/improve-prior-ship rules truncated ...]

## Action space is what you can verify exists

The mutation space is bounded by the runtime, the reachable web, and the
harness's current capability set. When a direction depends on something beyond
these -- a package, an API endpoint, a tool you assume is installed -- the
system treats unverified dependencies as hallucinations. You have `bash`,
`web_search`, and `web_fetch` to confirm a capability exists before writing code
against it; record the confirmation in `capability_evidence`.

## Build -> verify -> iterate (mandatory for code candidates)

For any candidate that introduces new executable code, you MUST complete this
loop IN YOUR SESSION before writing the manifest:

1. Write the code to your scratch dir.
2. Verify by actually running it -- not by reasoning about it. Two levels:
   - Level 1 -- unit call works: instantiate the processor/tool, drive the
     async hook, assert the expected state mutation happened.
   - Level 2 -- round-trip reaches the model: a unit call that returns does not
     prove the agent sees the return. Simulate the path from your code to the
     model's next input and assert the content survives it (provider serializer
     for tools; the next pipeline stage for processors).
3. Iterate if verification fails -- fix the bug, or pivot if the environment
   does not support what you assumed. Do NOT hide the failure in a try/except.
4. Attach the verifying output as `capability_evidence`. "I believe this will
   work" is not acceptable; paste the actual command and its output.

A candidate whose new code has not been observed to work will burn a round's
ship slot for zero flips. Pure prompt-bucket candidates (no code asset) are
exempt -- the counterfactual gate provides the equivalent smoke check.

[... reading list and write locations truncated ...]

## Manifest shape

Per candidate, emit a manifest at `{{ candidates_dir }}/C-R{{ round }}-<NN>.md`
and a scratch dir with the applied `config.yaml`.

---
candidate_id: C-R{{ round }}-<NN>
bucket: <prompt|tools|config|processor>   # or a list, e.g. [prompt, processor]
iterates_from: <prior_ship_id>            # OPTIONAL -- set for a revert/improve
capability_evidence:                      # REQUIRED -- may be empty []
  - type: <python_package|http_endpoint|builtin_tool|filesystem|other>
    claim: "<the capability this candidate depends on>"
    evidence: "<something you OBSERVED this session: command + output snippet>"
file_changes:
  - {path: <under scratch dir>, action: <create|modify|delete>, diff_summary: "<one line>"}
predicted_impact:
  tasks_will_unlock:    [<ALL_FAIL -> expect >=1 rollout to pass>]
  tasks_will_stabilize: [<PARTIAL_PASS -> expect all rollouts to pass>]
  tasks_at_risk:        [<currently >=1 pass -> might regress>]
attribution_signature:                    # recommended for tools/processor/config
  type: <tool_call|processor_invocation>
  tool_name: <PascalCase name as registered>
  expected_min_calls: 1
---

## Failure Evidence
At least one trajectory or digest anchor per candidate, e.g.
`trajectories/abc123_r0.jsonl#step_5 -- what went wrong here`.

## Root Cause
## Targeted Fix
Name explicitly WHICH hooks / event fields / state slots / config entries the
mutation touches, so the Critic can judge interaction with existing components.

## Why this won't break tasks_at_risk

[... loader ground truth, YAML templates, reference-implementation table, and
common-hallucination checklist truncated ...]
\end{promptfile}

\begin{promptfile}{critic/system\_prompt.md}
# Critic -- Round {{ round }}

Your goal has two parts, and both matter.

## Part 1 -- Per-candidate verdict

Pick the single candidate (or multiple bucket-disjoint candidates) whose
shipping is most likely to raise next round's pass rate without hurting it. If
none qualifies, no-op -- shipping a bad candidate is worse than nothing.

Every verdict MUST explicitly address candidate-vs-config interaction. Read the
candidate's `## Targeted Fix` (which hooks / event fields / state slots it
touches) AND the current HarnessConfig. Answer in your verdict: does this
candidate's mutation surface overlap with any processor, tool, prompt clause, or
config kwarg already in the parent config? If yes, argue whether the overlap is
(a) intentional and safe (the new component supersedes the old one, which the
candidate's applied YAML has removed) or (b) an accidental collision and grounds
for rejection. A verdict that does not address this is incomplete and counts as
ask-more.

[... round-trip (Level-2) evidence check for tool/processor candidates truncated ...]

## Part 2 -- Portfolio audit

Even when every individual candidate is acceptable, step back and look at the
pattern across rounds (scoreboard.json, data/ship_outcomes.json):

- For any lever item shipped in >=2 of the last 3 rounds with cumulative
  hit_rate < 0.4, do NOT ship a candidate touching that lever again; flag it as
  strategy_concern. A single-round miss is likely k-sampling noise; only
  persistence across rounds is signal.
- Is there a bucket or cluster the Evolver has never touched, while a failure
  pattern in digests/ suggests it is the right lever? Flag it.
- Did this round's regressions.md list any regressed task? The Evolver was
  required either to ship a candidate addressing each regression or to write a
  "## Why this regression is acceptable" section. Reject the round (no-op) if
  neither path was taken, citing the missed task IDs.

Record strategy_concern in decision.md's frontmatter only when the evidence is
concrete: name the bucket, the round range, the hit rate, the failing tasks.
Next round's Planner relays it to the Evolver. This is how you challenge the
Evolver's strategy, not just its candidates.

[... independence rule, available-to-read guide, ask_evolver, and loader ground
truth truncated ...]

## Output

For each candidate, write `verdicts/V-<candidate_id>.md`:

---
candidate_id: <C-R{{ round }}-NN>
verdict: <accept|reject|ask-more>
evidence_anchors:
  - trajectories/<file>#step_N
---

## Reasoning
<Why this verdict. Cite the anchors. 2-4 short paragraphs.>

After all verdicts, write `decision.md`:

---
round: {{ round }}
decision_type: <ship|no_op>
ship_ranking:                # candidates to ship, in priority order
  - candidate_id: <C-R{{ round }}-NN>
strategy_concern: |          # OPTIONAL -- fill only when the audit surfaces one
  <one concrete paragraph; cite ship_outcomes / task_history anchors>
---

## Reasoning
<3-6 bullets, one per verdict file, plus one bullet for any strategy_concern.>

Multi-ship: Stage 4 ships every listed candidate in order but skips any whose
bucket was already claimed by an earlier-ranked ship, so bucket-disjoint
candidates attacking orthogonal failure modes can ship together. Nothing ships
unless decision.md parses cleanly.
\end{promptfile}

\subsection{Round-0 Task-Agent Prompts}
\label{app:task-prompts}

The composition-layer default ($\mathcal{H}_0$) loads one system prompt per benchmark. We reproduce
the ALFWorld default below as a representative example; the remaining
benchmark defaults follow the same structure and are listed in the repository.

\begin{promptfile}{alfworld\_evolver/systemprompt.md}
# System

You are an expert agent operating in the ALFRED Embodied Environment. You drive
a live household simulator by calling the `act` tool, one admissible command per
call. The environment replies with the next observation and the current
admissible action list.

## Output Discipline

- One `act` call per turn. No chaining ("go to fridge 1 and open it").
- Pick the command verbatim from the admissible list. Anything outside that list
  is a silent no-op and wastes a step.
- Reply with `FINAL ANSWER: done` only after you see `__ALFWORLD_DONE__` or
  `__ALFWORLD_FAILED__` in an observation. Until then, keep calling `act`.
- If you genuinely cannot make progress for many consecutive steps, end with
  `FINAL ANSWER: give up`.

## Task Types

| Type            | Goal                          | Key Steps |
|-----------------|-------------------------------|-----------|
| Pick & Place    | Put object X in/on receptacle Y | Find X -> take X -> go to Y -> put X |
| Pick Two & Place| Put two instances of X in/on Y  | Find X1 -> take -> place -> find X2 -> take -> place |
| Examine in Light| Examine X under desklamp        | Find X -> take X -> find desklamp -> use desklamp |
| Clean & Place   | Clean X and put in/on Y         | Find X -> take -> clean at sinkbasin -> go to Y -> put |
| Heat & Place    | Heat X and put in/on Y          | Find X -> take -> heat at microwave -> go to Y -> put |
| Cool & Place    | Cool X and put in/on Y          | Find X -> take -> cool at fridge -> go to Y -> put |

## General Principles

1. Decompose the goal into ordered sub-goals (locate -> acquire -> transform ->
   deliver) and complete each before moving on.
2. Systematic exploration: search each surface and container at most once before
   revisiting. Open closed containers before judging them empty -- the
   admissible list surfaces `open <recep>` when you arrive at a closed one.
3. Grab immediately: when a required object appears, take it on the very next
   step before moving elsewhere.
4. Transform before placing: perform any clean/heat/cool state change at the
   appropriate appliance before heading to the final destination.
5. Direct delivery: once holding the goal object, navigate straight to the
   target receptacle and place it.
6. Track progress: keep an internal count of objects still to find and place.
   Only stop searching when the count reaches zero.
7. Avoid loops: never repeat the same action more than twice in a row. If stuck,
   move to a different unexplored location.
8. Trust the admissible list: if `take X from Y` does not appear, you are not at
   Y, Y is closed, or X is not visible -- `go to`, `open`, or move on rather
   than guessing.

## Common Mistakes to Avoid

- Revisiting searched locations without new evidence.
- Ignoring visible objects -- if the target appears, take it immediately.
- Skipping the state change -- do not place an object before cleaning / heating
  / cooling it when the task requires it.
- Premature termination -- do not reply `FINAL ANSWER: done` before the env
  emits `__ALFWORLD_DONE__`.
- Action loops -- repeatedly toggling or examining the same object wastes steps.
- Holding two objects at once -- you can only carry one. `put` the current one
  before `take`-ing the next.
\end{promptfile}

\subsection{Change-Manifest Schema}
\label{app:manifest-schema}

Each Evolver candidate is accompanied by a change manifest, a structured audit record linking the proposed edit to its evidence, mechanism, expected effect, and attribution signal. The manifest makes every harness modification falsifiable: the Critic checks whether the next round's trace features match the mechanism and impact the manifest predicted. Table~\ref{tab:manifest-fields} defines the manifest fields, and the schema below specifies their representation.

\begin{table}[!htbp]
\centering
\small
\caption{Change-manifest fields. The manifest is the loop's evidence ledger:
every shipped edit is falsifiable against the next round's trace-feature deltas.}
\label{tab:manifest-fields}
\begin{tabular}{lp{0.66\linewidth}}
\toprule
\textbf{Field} & \textbf{Meaning} \\
\midrule
candidate\_id        & Unique id, e.g.\ C-R3-01 (round 3, candidate 1). \\
bucket               & Edit type: prompt, tools, config, or processor. \\
capability\_evidence & Verified claims that the edit mechanism actually works. \\
file\_changes        & List of path / action / diff-summary edits. \\
predicted\_impact    & Tasks the edit will unlock, stabilize, or put at risk: the falsifiable prediction. \\
attribution\_signature & Trace feature that must appear if the edit fired, e.g.\ a processor invocation. \\
\bottomrule
\end{tabular}
\end{table}

\begin{manifestfile}{change\_manifest.yaml (schema)}
candidate_id: <round/candidate id>
bucket: prompt | tools | config | processor
capability_evidence:
  - {type: python_package | filesystem | other, claim: "...", evidence: "..."}
file_changes:
  - {path: "...", action: create | modify | delete, diff_summary: "..."}
predicted_impact:
  tasks_will_unlock:    [task_id, ...]
  tasks_will_stabilize: [task_id, ...]
  tasks_at_risk:        [task_id, ...]
attribution_signature:
  type: processor_invocation | tool_call | prompt_feature
  tool_name: <name>
  expected_min_calls: <int>
\end{manifestfile}

\section{Anatomy of an Evolution Step}
\label{app:anatomy}

To illustrate the AEGIS loop concretely, we walk through one full cycle, from Digester compression through Planner synthesis, Evolver editing, and Critic judgment, to the resulting trace delta. We select round~10 of the GAIA / Sonnet~4.6 run: a composite edit spanning a new tool, a prompt addition, and a configuration change. This multi-component intervention produced the largest single-round gain in that run, making it a richer illustration than a single-lever edit.

\subsection{Worked Example: GAIA / Sonnet 4.6, Round 10}
\label{app:worked-example}

\paragraph{Failure evidence.} Prior to round~10, the success rate stood at
$74.8\%$, having dropped from its peak of $77.7\%$ due to a regression in
round~9. The Digester's trace analysis revealed a systematic failure pattern: every Wikipedia fetch in round~10's traces returned zero characters. WebFetch employs a browser if the
website requires JavaScript support, but Wikipedia's new frontend fails to load
correctly, timing out or returning an empty body. The traces make it plain:
within task \texttt{db4fd70a} (number of stations in a rail line),
\texttt{db4fd70a\_r0.jsonl\#step\_0} and \texttt{\#step\_1} report that
Wikipedia WebFetch fetches return 0 chars; similarly, within
\texttt{f0f46385} (ASEAN members' membership status), three consecutive
WebFetch calls return 0 chars; ten separate attempts across the round returned empty responses.

\paragraph{Planner synthesis.} The Digester grouped the 23 failed tasks by failure mode, surfacing a critical tool-level issue that the round-9 Critic had already flagged: the \emph{tools} component had not shipped a fix in nine consecutive rounds despite source-access failures appearing since round~1. The Planner received two targets: (1)~resolve the persistent tool-level source-access failures, and (2)~revert the prompt and budget-processor changes responsible for the round-9 regression.

\paragraph{Evolver edit.} The Evolver suggested \texttt{C-R10-02}, covering three
buckets. (i)~\emph{tools}: new \texttt{WikiTextFetch} tool, avoiding the browser
altogether by employing the MediaWiki API endpoint; returns complete text of the
article, 10{,}529 chars in case of the rail line, 80{,}028 for ASEAN.
(ii)~\emph{prompt}: a single sentence in the tool usage section instructing the
agent to use \texttt{WikiTextFetch} before looking up Wikipedia articles. (iii)~\emph{config}:
restore the round-8 baseline configuration, register \texttt{WikiTextFetch},
and remove the problematic budget processor. The manifest's capability evidence includes a Level-2 round-trip check (content serialized as a 10{,}529-char string by the provider); the attribution signature requires at least one \texttt{WikiTextFetch} call.

\begin{manifestfile}{R10/candidates/C-R10-02.md}
candidate_id: C-R10-02
bucket: [tools, prompt, config]
capability_evidence:
  - type: http_endpoint
    claim: "MediaWiki API returns full plain-text extract where WebFetch returns 0 chars"
    evidence: "GET .../w/api.php?...&explaintext=true -> 10,529 chars for Franklin/Foxboro_Line"
  - type: other
    claim: "tool return survives provider serialization to the model (Level 2)"
    evidence: "_prepare_messages([tool_msg]) keeps content as 10,529-char string"
file_changes:
  - {path: R10/applied/C-R10-02/wiki_text_fetch.py, action: create, diff_summary: "WikiTextFetch via MediaWiki API"}
  - {path: R10/applied/C-R10-02/gaia_agent.md,     action: create, diff_summary: "R8 prompt + one WikiTextFetch line"}
  - {path: R10/applied/C-R10-02/config.yaml,        action: create, diff_summary: "register tool; restore R8; drop budget processor"}
predicted_impact:
  tasks_will_unlock:    [db4fd70a, f0f46385, 983bba7c, 08f3a05f, 5e2a91b0]
  tasks_will_stabilize: [4b6bb5f7, 42d4198c]
  tasks_at_risk:        []
attribution_signature:
  type: tool_call
  tool_name: WikiTextFetch
  expected_min_calls: 1
\end{manifestfile}

\paragraph{Critic verdict.} The Critic approved only \texttt{C-R10-02}, rejecting the competing revert-only candidate \texttt{C-R10-01}. Its rationale covered three aspects. (i)~\emph{Interaction}: \texttt{C-R10-02} is a strict superset of \texttt{C-R10-01}; both restore the round-8 baseline, so the budget-processor removal is intentional, not an accidental overlap. (ii)~\emph{Round-trip evidence}: the Critic verified Level-2 evidence (tool output arrives as a full string, not a truncation marker) before accepting any tools-bucket candidate. (iii)~\emph{Portfolio}: this is the first tools-bucket ship in ten rounds, and the trace evidence of persistent zero-char returns justifies the intervention that round-9 flagged.

\paragraph{Delta realized.} Post-shipping, the GAIA pass rate increased from
$74.8\%$ at R9 to $79.6\%$ at R10 (+$4.9$pp, five tasks changed to pass), the
greatest improvement during the entire run; five of the seven tasks the
tool was predicted to affect flipped to pass (hit rate $0.71$, the highest for any
ship across all 19 runs). The improvement mainly occurred at Levels~2 (+$4$
tasks) and~3 (+$2$). Since the tool triggered its target tasks, the attribution
condition was satisfied.

Figure~\ref{fig:c-r10-02-card} summarizes the same edit as a manifest card: the
raw YAML above is what the loop logs, the card is the human-facing reading of it.

\begin{figure}[!htbp]
\centering
\begin{manifestcard}{\texttt{C-R10-02}\, $\vert$\, GAIA / Sonnet 4.6\, $\vert$\, round 10\, $\vert$\, bucket: tools+prompt+config}
\mfsec{Failure evidence}
Every Wikipedia WebFetch in the round's trajectories returns 0 chars (10+ calls
across \texttt{db4fd70a}, \texttt{f0f46385}, \texttt{42d4198c}); the headless
browser fallback times out on Wikipedia's frontend.
\mfsec{Edit}
New \texttt{WikiTextFetch} tool calls the MediaWiki API for plain-text article
content (tools); one prompt line routes Wikipedia lookups to it first (prompt);
the round-8 baseline is restored and the regressing budget processor removed
(config).
\mfsec{Predicted fixes}
Unlock \texttt{db4fd70a}, \texttt{f0f46385}, \texttt{983bba7c},
\texttt{08f3a05f}, \texttt{5e2a91b0}; stabilize \texttt{4b6bb5f7}, \texttt{42d4198c}.
\mfsec{Attribution signature}
\texttt{tool\_call} on \texttt{WikiTextFetch}, $\geq 1$ call; verified against
the next round's trace.
\end{manifestcard}
\caption{Change manifest for \texttt{C-R10-02} rendered as a manifest card, the
human-facing counterpart of the logged YAML above.}
\label{fig:c-r10-02-card}
\end{figure}

\section{Additional Results}
\label{app:additional}

The rest of this appendix is organized per benchmark. Each subsection is built
around one figure of three panels: \textbf{(a)} a breakdown of the failure
clusters the adaptation loop had to address, \textbf{(b)} the per-model
distribution of harness levers the evolution shipped, and \textbf{(c)} a
model-by-lever heatmap of each lever's effectiveness (tasks flipped to pass over
tasks predicted). We read the three panels in order: what fails and why, how
each model evolves, and whether the evolution closes the failures. 

\subsection{GAIA}
\label{app:gaia-detail}

\begin{figure*}[!htbp]
\centering
\includegraphics[width=\linewidth]{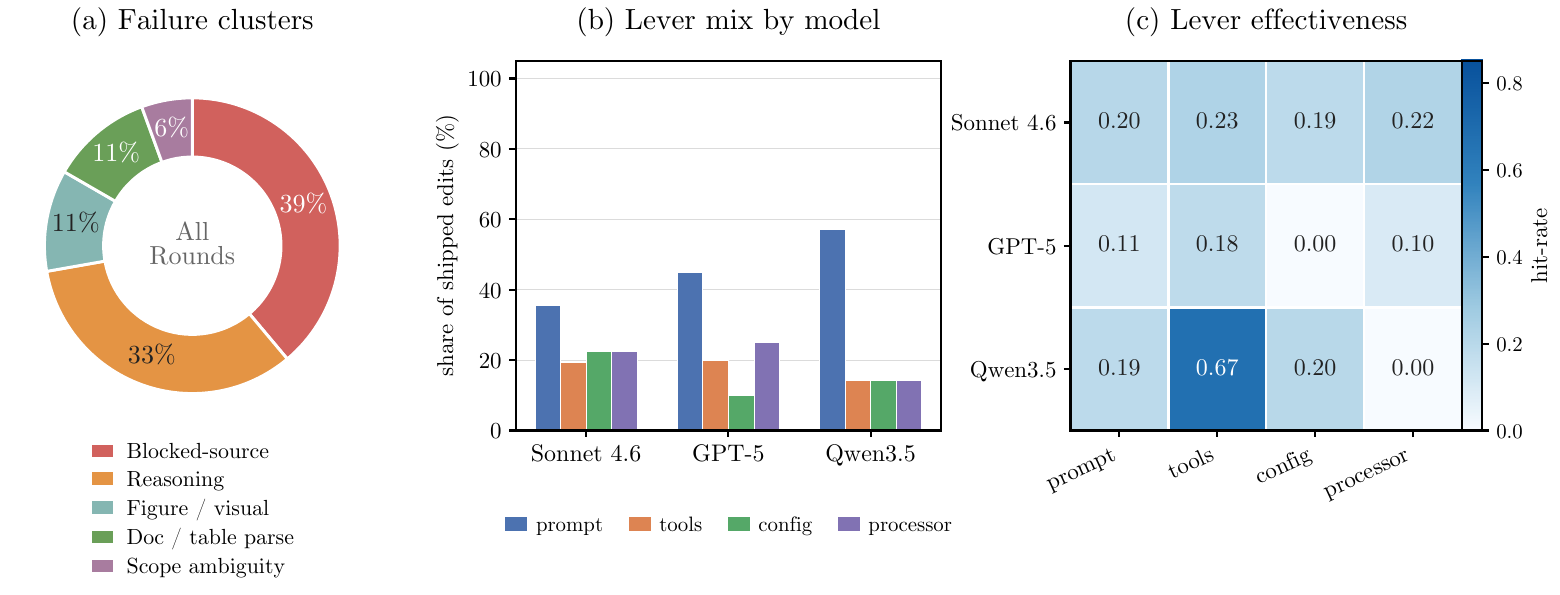}
\caption{GAIA evolution analysis (103 tasks, exact-match). \textbf{(a)}~Failure
clusters among the tasks still unsolved; blocked-source and reasoning dominate, while figure/visual and parsing clusters are residual model gaps. \textbf{(b)}~Share of
shipped edits by bucket for each task model. \textbf{(c)}~Lever effectiveness as hit-rate (tasks flipped / predicted) per model and bucket. The single Qwen3.5 tools ship is the highest-yield cell ($0.67$).}
\label{fig:gaia-detail}
\end{figure*}

GAIA stresses general reasoning under tool use, and is the most lever-diverse
benchmark in our suite. Figure~\ref{fig:gaia-detail} gives the three views we
use throughout this appendix.

\paragraph{Failure clusters and their causes.} Panel~(a) summarizes the failure clusters accumulated across the GAIA run. The dominant cluster is blocked-source ($39\%$), where the agent cannot retrieve the required evidence because pages return empty content, require JavaScript rendering that times out, or contain incomplete information. Reasoning failures ($33\%$) follow, covering tasks that require multi-hop inference, disambiguation of similar entities, numerical computation, or precise interpretation of underspecified queries. Figure/visual failures ($11\%$) arise when the answer depends on information embedded in images, maps, or diagrams that text extraction alone cannot capture. Document/table parsing failures ($11\%$) occur when evidence is locked in PDFs, structured tables, or semi-structured formats and the agent either misparses the layout or overlooks relevant cells. Scope ambiguity ($6\%$) covers queries with multiple valid interpretations, where the agent answers a related but incorrect reading. Together, these clusters indicate that GAIA failures concentrate in evidence retrieval, multi-step reasoning, visual grounding, structured-document extraction, and query disambiguation.

\paragraph{Per-model evolution logic.} Panel~(b) shows GAIA is the only
benchmark where all four levers see substantial use; the Sonnet run alone
shipped 11 prompt, 7 processor, 7 config, and 6 tools edits, because its failure
set spans tool, prompt, and config problems simultaneously. The three models
nonetheless diverge in a way that tracks their starting competence: Sonnet, with
the most rounds, sweeps every lever; GPT-5.4 leans hardest on prompt ($45\%$ of its
ships) and barely touches config, since its reasoning is already strong enough
that the remaining gains are mostly instruction-following; Qwen3.5's short run
concentrates its few ships and, strikingly, lands its single tools ship at the
highest yield of any cell. The shared logic is prompt-first for the behavioral
clusters, with the scarce tools lever reserved for the one mechanical cluster
prose cannot touch.

\paragraph{Did evolution close the clusters?} Panel~(c) shows which failure clusters were reduced by evolution. The largest improvement comes from the blocked-source cluster: the tool edit that introduced \texttt{WikiTextFetch} replaced unreliable browser-based Wikipedia fetching with a MediaWiki API call, reducing failures caused by empty or incomplete page retrieval. Prompt edits mainly targeted the reasoning cluster by encouraging more explicit verification, which contributed to steady gains across rounds. By contrast, figure / visual and document / table parsing failures remained harder to reduce, because they require information extraction from images, figures, PDFs, or structured tables. Overall, GAIA improves through a combination of tool edits that fix retrieval failures and prompt edits that reduce reasoning errors, while residual errors concentrate in visual and document-heavy tasks.

\subsection{ALFWorld}
\label{app:alfworld-detail}

\begin{figure*}[!htbp]
\centering
\includegraphics[width=\linewidth]{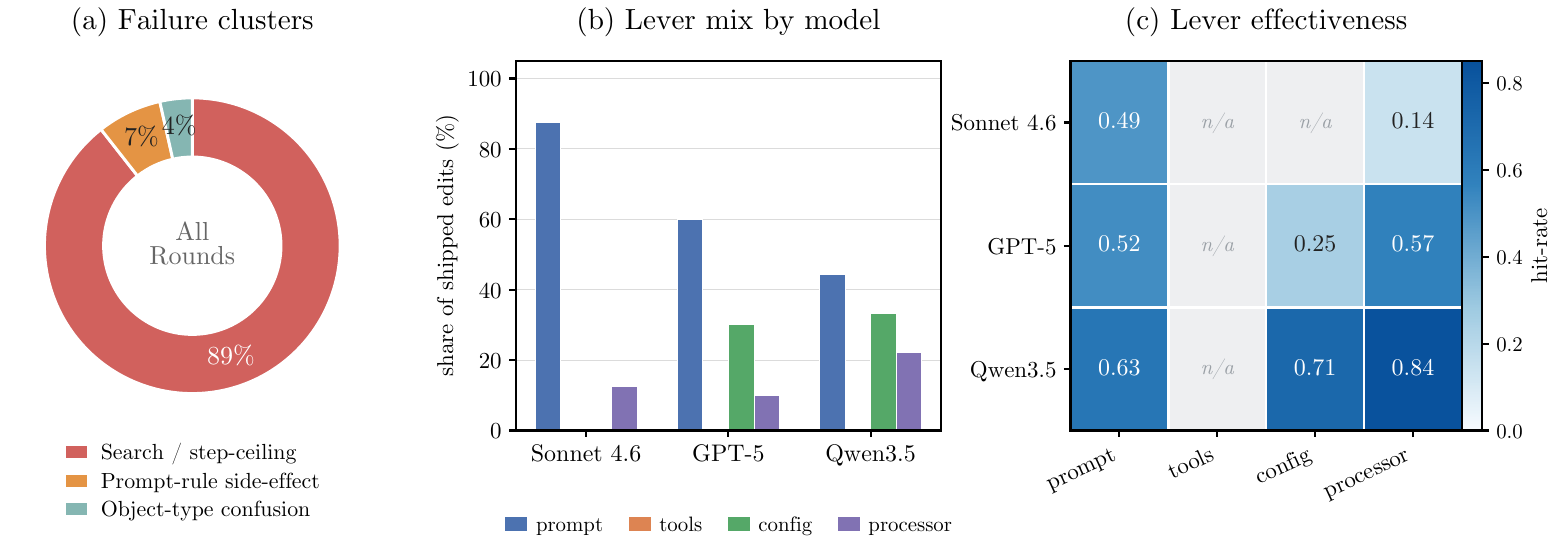}
\caption{ALFWorld evolution analysis (134 tasks, goal-completion).
\textbf{(a)}~Failure clusters accumulated across all rounds; search inefficiency
and the hard step-ceiling dominate, with two small clusters that evolution itself
introduced (a prompt-rule side-effect) or transiently hit (object-type
confusion). \textbf{(b)}~Lever mix by model: the strong base (Sonnet) climbs on
prompt almost alone, while weaker bases reach for more varied levers.
\textbf{(c)}~Lever effectiveness: structural levers (processor, config) are both
used more and more effective on weaker models.}
\label{fig:alfworld-detail}
\end{figure*}

ALFWorld is an embodied planning benchmark and the most prompt-dominated in our
suite. Figure~\ref{fig:alfworld-detail} shows its clusters, per-model logic, and
effectiveness.

\paragraph{Failure clusters and their causes.} Panel~(a) summarizes the main failure clusters observed on ALFWorld. The dominant cluster is search / step-ceiling ($89\%$), which covers episodes where the agent either searches rooms or receptacles in an inefficient order, or reaches the step limit before completing long interaction chains such as deep search or transform-then-place tasks. Prompt-rule side-effect failures ($7\%$) occur when an added heuristic improves some tasks but unintentionally restricts behavior on others, causing the agent to skip a valid action or stop searching too early. Object-type confusion failures ($4\%$) refer to cases where the agent confuses semantically similar objects. Together, these clusters show that ALFWorld failures mainly arise from search efficiency, over-constrained prompting, and object-specific grounding errors.

\paragraph{Per-model evolution logic.} Panel~(b) shows that prompt dominance scales inversely with base-model strength: a prompt rule yields gains only for models that reliably follow it. Sonnet, being the strongest base, derives nearly all improvement from prompt edits alone; search-order heuristics in the system prompt suffice because it consistently obeys them. GPT-5.4 supplements prompts with a processor (introduced at round three) that manages its step budget for transformation tasks. Qwen3.5 requires the most varied mix (prompt, config, and processor), including a processor that intercepts its reasoning text and re-emits tool calls when needed, a mechanical fix for a failure that prompt-level steering cannot resolve. The shared pattern is prompt-first, with structural levers recruited only when prompts prove insufficient. The weaker the base model, the sooner evolution falls back from prompt-based steering to config or processor enforcement, visible in the growing non-prompt segments from Sonnet to Qwen.

\paragraph{Did evolution close the clusters?}
Panel~(c) shows that evolution reduced the main ALFWorld failure clusters, with different levers mattering for different task agents. For Qwen3.5, processor and config edits achieved the strongest effects, with hit-rates of $0.84$ and $0.71$, respectively. These edits directly addressed mechanical failures by re-emitting missed tool calls and adjusting execution budgets, allowing Qwen3.5 to improve by $+44.0$pp and approach the closed-model runs. For Sonnet, the remaining failures were less structural, so prompt edits were sufficient for most gains, reaching a $0.49$ hit-rate, while the processor edit had only a marginal effect ($0.14$). Two clusters were only partially reduced: prompt-rule side effects, which were introduced by some evolved heuristics and then patched in later rounds, and long-path failures, where some episodes still exceeded the available interaction budget. Overall, ALFWorld shows a clear model-dependent pattern: stronger models benefit mainly from prompt-level steering, whereas weaker models require more structural support through processor and configuration edits.

\subsection{WebShop}
\label{app:webshop-detail}

\begin{figure*}[!htbp]
\centering
\includegraphics[width=\linewidth]{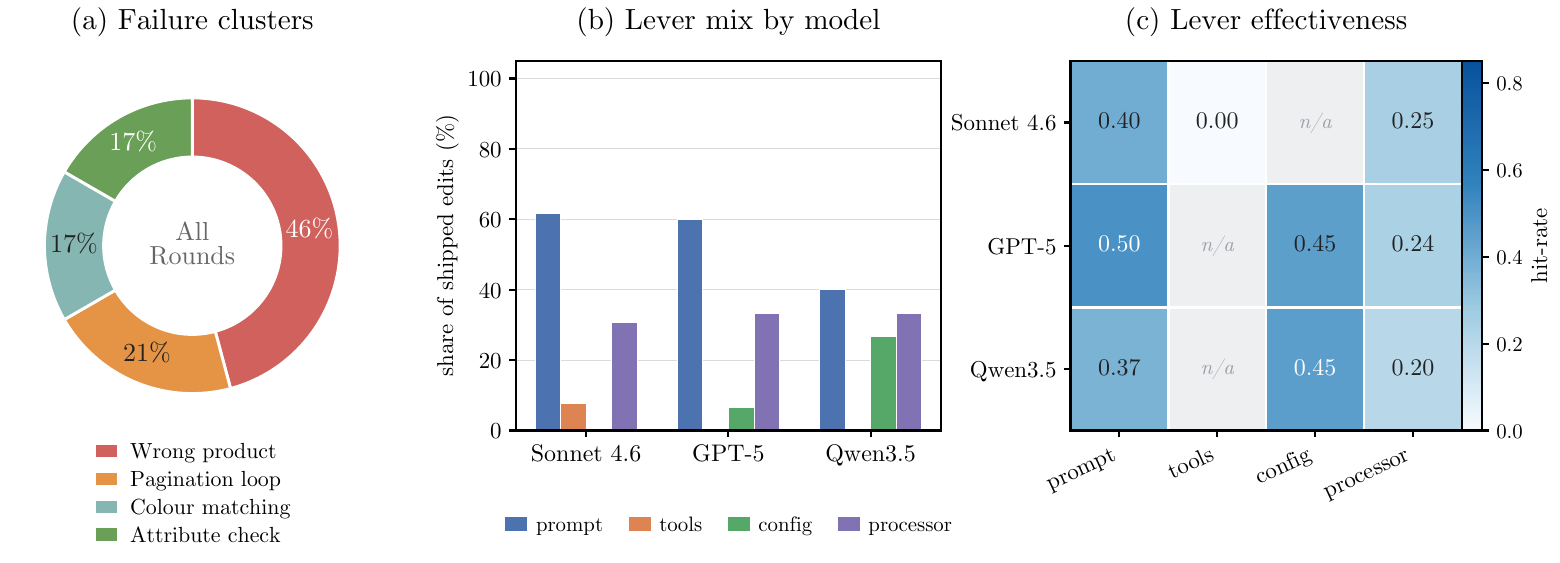}
\caption{WebShop evolution analysis (100 sessions).
\textbf{(a)}~Failure clusters across the run, after evolution has tamed the
round-0 search/pagination loops; the residual is mostly product-choice judgment
(wrong product, colour matching, attribute check). \textbf{(b)}~Lever mix by
model: prompt carries the climb, processor is the consistent second lever.
\textbf{(c)}~Lever effectiveness; prompt and config are the productive levers,
the lone colour-matcher tool ship returned $0.0$.}
\label{fig:webshop-detail}
\end{figure*}

WebShop is a web-interaction benchmark and the noisiest run in our suite.
Figure~\ref{fig:webshop-detail} shows its clusters, per-model logic, and
effectiveness.

\paragraph{Failure clusters and their causes.}
Panel~(a) summarizes the WebShop failure clusters accumulated across the run. Early failures are dominated by search and pagination loops, where the agent repeatedly reformulates queries or cycles through result pages without committing to a purchase. As evolution reduces these control-flow errors, the remaining failures shift toward product-selection judgment. The largest cluster, wrong product ($46\%$), occurs when the agent selects an item from the wrong category or settles on a weak match before comparing alternatives. Pagination loop failures ($21\%$) capture the remaining cases of repeated next/previous navigation without progress. Colour matching failures ($17\%$) arise when the agent mishandles shade equivalence or site-specific color labels, such as treating ``wine'' and ``red'' as incompatible. Attribute check failures ($17\%$) occur when the selected item is close to the request but fails on a required detail, such as size, sleeve length, or another unverified attribute. Overall, the cluster shift indicates that evolution first reduces navigation loops, after which the main errors concentrate in product matching and attribute verification.

\paragraph{Per-model evolution logic.}
Panel~(b) shows that prompt edits drive most of the improvement across all three models, with processor edits serving as a consistent secondary lever. This pattern matches WebShop's main control-flow failures. Prompt rules help the agent search more efficiently and commit earlier, while advisory processors reinforce these rules during execution by adding warnings when the agent begins to repeat searches or cycle through pagination. For product-selection failures, evolution introduces more targeted support: a colour-matching tool helps resolve shade-equivalence cases, and config edits help weaker models maintain context over longer shopping sessions. Overall, WebShop requires a mixed response: prompts improve high-level shopping strategy, processors reduce navigation loops, tools support attribute matching, and config edits stabilize long-session behavior.

\paragraph{Did evolution close the clusters?}
Panel~(c) shows that evolution partially reduced the WebShop failure clusters. Prompt edits were the most consistently effective lever across models, with hit-rates of $0.37$--$0.50$, while config edits helped the two weaker models maintain context over longer sessions. These changes reduced early search and pagination loops, raising performance from $60\%$ to a peak of $76\%$. The remaining clusters proved harder to close. Advisory processors produced only modest gains ($0.20$--$0.25$), so some pagination failures persisted. The colour-matching tool did not improve performance in this run ($0.0$ hit-rate), leaving the colour-matching cluster largely unchanged. Overall, WebShop benefits most from prompt and config edits, while residual navigation loops and product-judgment errors remain the main sources of instability.

\subsection{$\tau^3$-Bench}
\label{app:tau3-detail}

\begin{figure*}[!htbp]
\centering
\includegraphics[width=\linewidth]{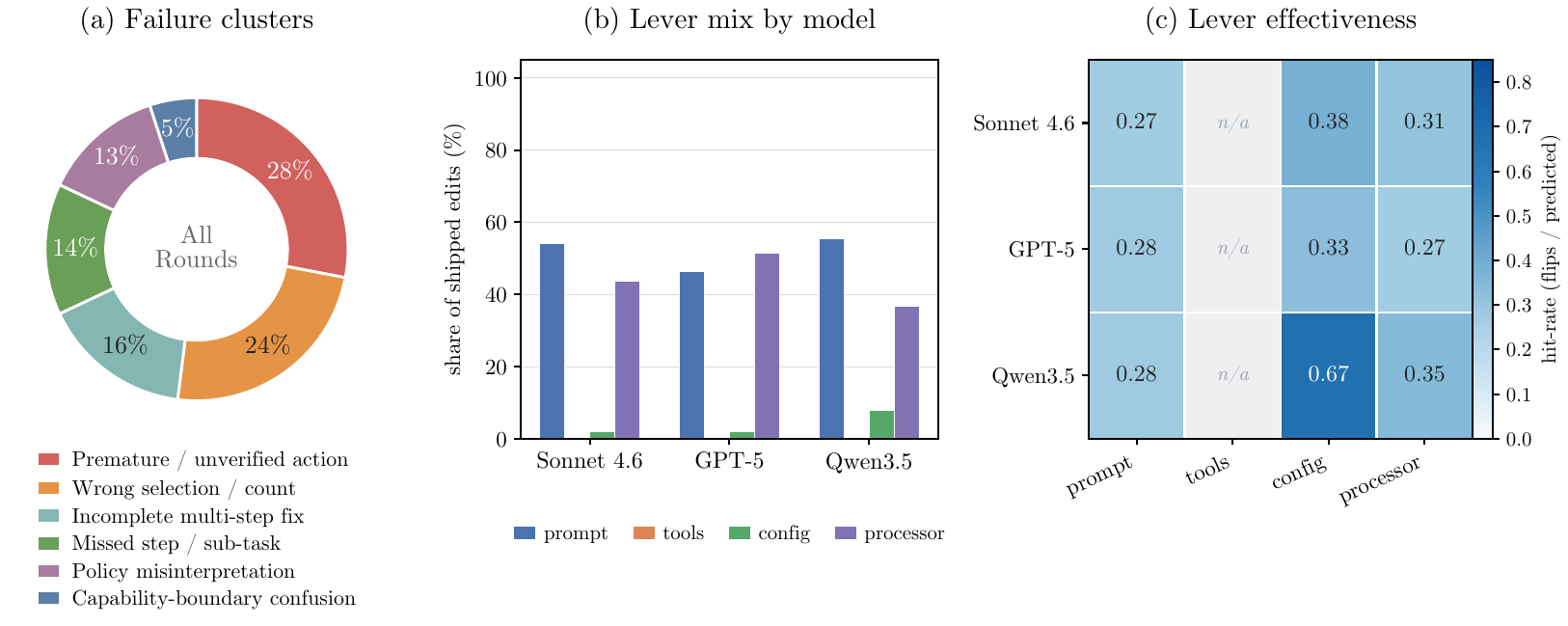}
\caption{$\tau^3$-Bench evolution analysis, pooled over the airline, retail, and
telecom domains. \textbf{(a)}~Failure clusters from the logged digests
(harness-interrupt traces excluded); judgment errors (premature action and
wrong selection) dominate. \textbf{(b)}~Lever mix by model: prompt and
processor carry the climb, with zero tools edits since the tool set is fixed.
\textbf{(c)}~Lever effectiveness; config is sharpest where used (Qwen3.5 $0.67$)
but rare, while prompt and processor are the consistent high-volume levers.}
\label{fig:tau3-detail}
\end{figure*}

$\tau^3$-Bench stresses multi-turn dialogue under an explicit domain policy.
Figure~\ref{fig:tau3-detail} pools the AEGIS runs across the airline, retail,
and telecom domains.

\paragraph{Failure clusters and their causes.}
Excluding harness-interrupt traces, the failures are judgment-heavy: the top two
clusters, premature / unverified action ($28\%$; committing a booking, refund,
or device fix before a precondition holds) and wrong selection / count
($24\%$), together exceed half and concern \emph{when} to commit and
\emph{what} to pick rather than mechanical execution. The remainder is
procedural (incomplete multi-step fix $16\%$, missed step / sub-task $14\%$) or
policy-related (misinterpretation $13\%$). The smallest cluster,
capability-boundary confusion ($5\%$), is $\tau^3$-specific: some telecom faults
live on the user's handset, where the agent has no device-side tool and the
failure is its treating that boundary as a missing capability.

\paragraph{Per-model evolution logic.}
Evolution is prompt-and-processor driven for every model, with zero tools edits:
the tool set is fixed and no cluster is one that a new tool could close. Sonnet 4.6
splits prompt/processor ($23/18$), GPT-5.4 ships the most balanced mix
($19/20$), and Qwen3.5-9B ships fewer overall ($14/9$). Since $\tau^3$ failures
are control-flow and judgment errors, the productive levers are prompt rules
that encode the policy's ordering constraints and processors that enforce them
mid-dialogue.

\paragraph{Did evolution close the clusters?}
Config is the sharpest lever where used (Qwen3.5 $0.67$, GPT-5.4 $0.33$ hit-rate)
but is shipped rarely; the high-volume prompt and processor levers are moderately
effective ($0.27$--$0.35$), matching the control-flow nature of the dominant
clusters. Gains track base-model headroom: GPT-5.4 starts lowest ($76.2\%$) and
gains most ($+14.5$pp), Sonnet 4.6 gains $+5.4$pp, and near-ceiling Qwen3.5-9B
only $+1.1$pp. The loop is not monotone; Sonnet's telecom run reaches $100\%$
at R4, regresses to $80.7\%$ at R7 after a sixth consecutive same-bucket edit,
then recovers to $99.1\%$ by R9 (Section~\ref{sec:exp-failure}). Overall, the
ordering-enforcing levers close the premature-action and missed-step clusters
most reliably, while wrong-selection and policy-judgment errors are the harder
residual.

\subsection{SWE-bench Verified}
\label{app:swe-detail}

\begin{figure*}[!htbp]
\centering
\includegraphics[width=\linewidth]{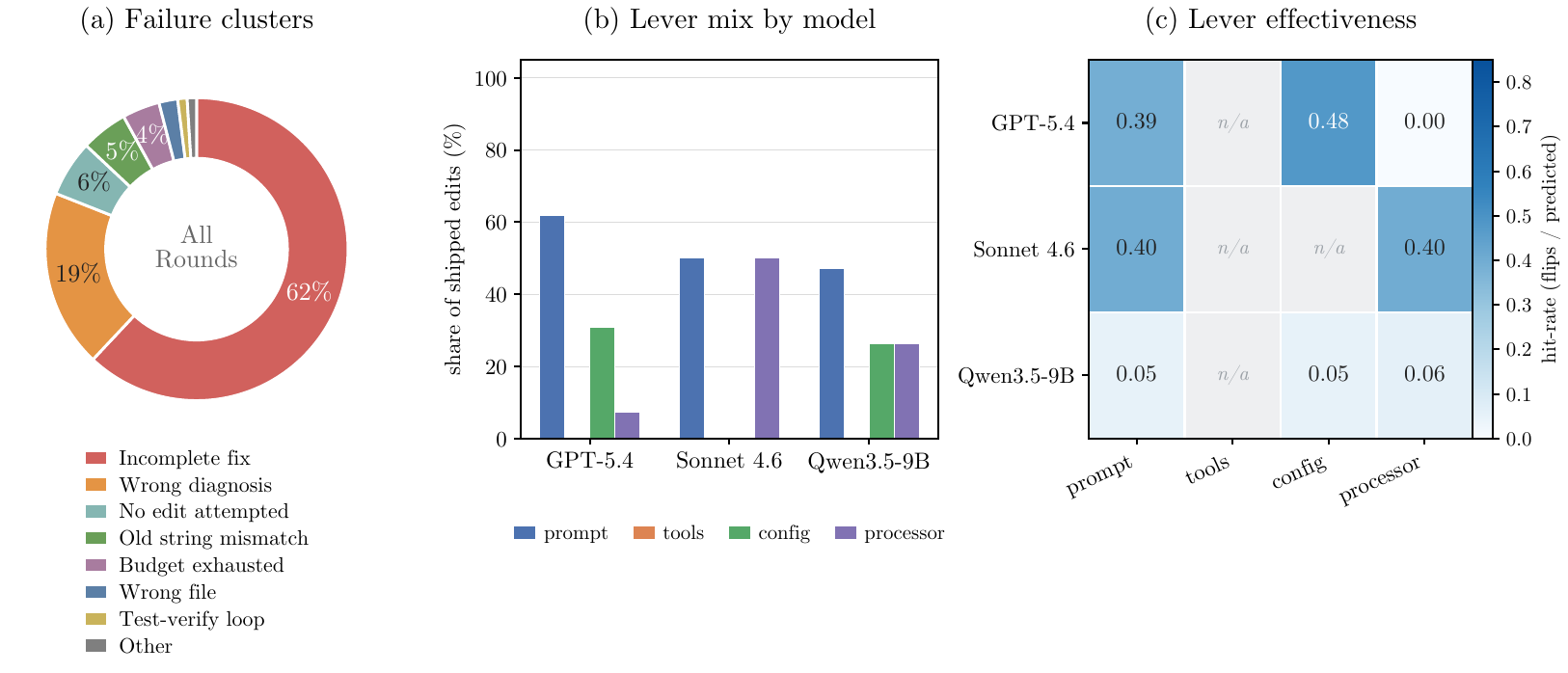}
\caption{SWE-bench Verified evolution analysis ($55$ tasks, resolved-rate).
\textbf{(a)}~Failure clusters pooled across all rounds and all three task models;
incomplete fix and wrong diagnosis dominate, while the mechanical tail (no-edit,
anchor mismatch, budget) is residual; failures are under-fixes, not gamed
evaluations. \textbf{(b)}~Lever mix by model: every run is prompt-first and ships
zero tools edits, with the secondary lever shifting from processor (Sonnet) to
config (GPT-5.4) to a varied mix (Qwen3.5). \textbf{(c)}~Lever effectiveness as
hit-rate (tasks flipped / predicted); strong models reach $0.39$--$0.48$ on their
productive levers, whereas every Qwen3.5-9B lever collapses to $\approx0.05$, a
capability floor below which evolution cannot compound.}
\label{fig:swe-detail}
\end{figure*}

SWE-bench Verified stresses repository-level code editing. 
Figure~\ref{fig:swe-detail} shows its clusters, per-model logic, and effectiveness.

\paragraph{Failure clusters and their causes.} 

Panel~(a) summarizes the failure clusters pooled across all rounds and all three models. The dominant cluster is incomplete fix ($62\%$), where the agent reaches the right region and produces a valid patch but covers only one branch or call site while the gold patch needs several. Wrong diagnosis ($19\%$) follows, covering edits to the wrong file or abstraction level after misreading the root cause. The remaining tail is mechanical rather than cognitive: no edit attempted ($6\%$), \texttt{Edit} anchor mismatch ($5\%$), and budget exhausted ($4\%$). Notably the composition is the inverse of reward-hacking: failures are under-fixes, not gamed evaluations, because the harness applies the gold test patch before the model patch and blocks test-file writes.
  
\paragraph{Per-model evolution logic.} 

Panel~(b) shows that SWE-bench is prompt-first for every model, with the secondary lever tracking base-model strength. All three runs ship zero tools edits, since unlike GAIA the failure set has no mechanical-retrieval cluster a tool could close. Sonnet pairs prompt with an equal share of processor edits ($7$ each), using workflow nudges to shape an already-competent coder; GPT-5.4 leans hardest on prompt ($8$ ships) and uses config ($4$) to revert a harmful nudge and restructure its strategy phases; Qwen3.5 spreads its few ships across prompt, processor, and config ($6/3/3$). The shared logic is prompt-first, with structural levers recruited as the base model weakens.

\paragraph{Did evolution close the clusters?} 
Panel~(c) reveals a sharp capability floor. For the strong models the productive levers genuinely close failures: GPT-5.4's config edits reach $0.48$ and prompt $0.39$, while Sonnet's prompt and processor levers both land at $0.40$. For Qwen3.5-9B every lever collapses to near-zero (prompt $0.05$, config $0.05$, processor $0.06$), an order of magnitude lower, because the $9$B base cannot execute the predicted fixes. The same loop that lifts GPT-5.4 from $45\%$ to a $64\%$ peak and stabilizes Sonnet near $87\%$ yields only noise on Qwen3.5 (peak $42\%$, zero durable gains). Overall SWE-bench improves through prompt edits that broaden fix scope and config edits that restore workflow pacing, but only for models strong enough to act on them.

\section{Reproducibility and Artifacts}
\label{app:artifacts}

\subsection{Per-Run Directory Layout}
Each evolution run writes a self-describing directory. The layout below lets a
reader reconstruct any decision in this paper from the logged artifacts.

\begin{promptfile}{runs/<run\_name>/ (per-run artifact layout)}
runs/<run_name>/
|-- INDEX.md            # human-readable index of the run
|-- journal.md          # first-person memo, one entry per round
|-- curves.json         # per-round pass-rate trajectory
|-- scoreboard.json     # ships + per-bucket reputation
|-- audit.jsonl         # structured event log (stage / gate / commit)
|-- data/
|   |-- task_history.jsonl      # one line per (round, task)
|   |-- ship_outcomes.json      # one entry per historical ship
|   `-- rejected_candidates.jsonl
`-- R<n>/                        # per-round artifacts
    |-- landscape.md             # Planner cross-trace synthesis
    |-- candidates/C-R<n>-NN.md  # Evolver change manifests (K per round)
    |-- applied/C-R<n>-NN/       # applied config + asset files
    |-- decision.md              # Critic ship / no_op decision
    |-- verdicts/V-C-R<n>-NN.md  # per-candidate verdicts
    |-- regressions.md           # tasks worsened vs R<n-1>
    |-- digests/*.md             # per-task failure analysis
    `-- trajectories/*.jsonl     # raw rollouts
\end{promptfile}

\addtocontents{toc}{\protect\setcounter{tocdepth}{2}}

\end{document}